\documentclass[journal]{IEEEtran}

\usepackage{xcolor,soul,framed} 

\colorlet{shadecolor}{yellow}
\usepackage[pdftex]{graphicx}
\graphicspath{{../pdf/}{../jpeg/}}
\DeclareGraphicsExtensions{.pdf,.jpeg,.png}

\usepackage[cmex10]{amsmath}
\usepackage{array}
\usepackage{hyperref}
\usepackage{mdwmath}
\usepackage{longtable}
\usepackage{tabularx}
\usepackage{mdwtab}
\usepackage{eqparbox}
\usepackage{url}
\usepackage{graphicx}
\usepackage[center]{caption}
\hypersetup{
    colorlinks=true,
    linkcolor=red,
    filecolor=magenta,      
    urlcolor=blue,
    citecolor=red,
}

\begin{document}
\bstctlcite{IEEEexample:BSTcontrol}
    \title{A Review of Deep Learning with Special Emphasis on Architectures, Applications and Recent Trends}
  \author{Saptarshi Sengupta, Sanchita Basak, Pallabi Saikia, Sayak Paul, Vasilios Tsalavoutis, Frederick Ditliac Atiah, Vadlamani Ravi and Richard Alan Peters II

  \thanks{Manuscript received March XX, 2019.}
  \thanks{S. Sengupta, S. Basak and R.A. Peters II are with the EECS Department at Vanderbilt University, USA. {e-mail: saptarshi.sengupta@vanderbilt.edu; sanchita.basak@vanderbilt.edu; alan.peters@vanderbilt.edu}}
  \thanks{P. Saikia is with the CSE Department at IIT Guwahati, India. {e-mail:  pallabi.s@iitg.ac.in}}
  \thanks{Sayak Paul is with DataCamp Inc, India. {e-mail: spsayakpaul@gmail.com}}
  \thanks{V. Tsalavoutis is with the School of Mechanical Engineering, Industrial Engineering Laboratory, National Technical University of Athens, Greece. {e-mail: btsalavoutis@mail.ntua.gr}}
   \thanks{F. Atiah is with the Computer Science Department at the  University of Pretoria. {e-mail: u16403381@tuks.co.za}}
\thanks{R. Vadlamani is with the Institute for Development and Research in Banking Technology, India. {e-mail:  vravi@idrbt.ac.in}}}
  
\markboth{IEEE TRANSACTIONS ON XXX, VOL.~XX, NO.~YY, March~2019
}{Sengupta \MakeLowercase{\textit{et al.}}: A Survey of Deep Neural Networks with Applications and Future Directions}

\maketitle
\begin{abstract}

Deep learning has solved a problem that as little as five years ago was thought by many to be intractable - the automatic recognition of patterns in data; and it can do so with an accuracy that often surpasses that of human beings. It has solved problems beyond the realm of traditional, hand-crafted machine learning algorithms and captured the imagination of practitioners trying to make sense out of the flood of data that now inundates our society. As public awareness of the efficacy of deep learning increases so does the desire to make use of it. But even for highly trained professionals it can be daunting to approach the rapidly increasing body of knowledge produced by experts in the field. Where does one start? How does one determine if a particular Deep Learning model is applicable to their problem? How does one train and deploy such a network? A primer on the subject can be a good place to start. With that in mind, we present an overview of some of the key multilayer artificial neural networks that comprise deep learning. We also discuss some new automatic architecture optimization protocols that use multi-agent approaches. Further, since guaranteeing system uptime is becoming critical to many computer applications, we include a section on using neural networks for fault detection and subsequent mitigation. This is followed by an exploratory survey of several application areas where deep learning has emerged as a game-changing technology: anomalous behavior detection in financial applications or in financial time-series forecasting, predictive and prescriptive analytics, medical image processing and analysis and power systems research. The thrust of this review is to outline emerging areas of application-oriented research within the deep learning community as well as to provide a handy reference to researchers seeking to use deep learning in their work for what it does best: statistical pattern recognition with unparalleled learning capacity with the ability to scale with information.
\end{abstract}


\IEEEpeerreviewmaketitle


\section{Introuction}

Artificial neural networks (ANNs), now one of the most widely-used approaches to computational intelligence, started as an attempt to mimic adaptive biological nervous systems in software and customized hardware \cite{10.3389/fncom.2017.00114}. ANNs have been studied for more than 70 years \cite{McCulloch1943} during which time they have waxed and waned in the attention of researchers. Recently they have made a strong resurgence as pattern recognition tools following pioneering work by a number of researchers \cite{lecun2015deep}. It has been demonstrated unequivocally that multilayered artificial neural architectures can learn complex, non-linear functional mappings, given sufficient computational resources and training data. Importantly, unlike more traditional approaches, their results scale with training data. Following these remarkable, significant results in robust pattern recognition, the intellectual neighborhood has seen exponential growth, both in terms of academic and industrial research. Moreover, multilayer ANNs reduce much of the manual work that until now has been needed to set up classical pattern recognizers. They are, in effect, black box systems that can deliver, with minimal human attention, excellent performance in applications that require insights from unstructured, high-dimensional data \cite{554195} \cite{7298965} \cite{Donahue:2017:LRC:3069214.3069251} \cite{DBLP:journals/corr/WuHS15} \cite{8100028} \cite{7506134}. These facts motivate this review of the topic.


\subsection {\textbf{What is an Artificial Neural Network?}}

An artificial neural network comprises many interconnected, simple functional units, or \textit{neurons} that act in concert as parallel information-processors, to solve classification or regression problems.
That is they can separate the input space (the range of all possible values of the inputs) into a discrete number of classes or they can approximate the function (the black box) that maps inputs to outputs.
If the network is created by stacking layers of these multiply connected neurons the resulting computational system can:
\vspace{1 em}
\begin{enumerate}
    \item Interact with the surrounding environment by using one layer of neurons to receive information (these units are known to be part of the \textit {input layers} of the neural network)
    \vspace{1 em}
    \item Pass information back-and-forth between layers within the black-box for processing by invoking certain \textit{design goals} and \textit{learning rules} (these units are known to be part of the \textit {hidden layers} of the neural network)
    \vspace{1 em}
    \item Relay processed information out to the surrounding environment via some of its atomic units (these units are known to be part of the \textit {output layers} of the \textit {neural network}).
    \vspace{1 em}
    \end{enumerate}
    
Within a hidden layer each neuron is connected to the outputs of a subset (or all) of the neurons in the previous layer each multiplied by a number called a \textit{weight}.
The neuron computes the sum of the products of those outputs (its inputs) and their corresponding weights.
This computation is the dot product between an input vector and weight vector which can be thought of as the projection of one vector onto the other or as a measure of similarity between the
the two.
Assume the input vectors and weights are both $n$-dimensional and there are $m$ neurons in the layer.
Each neuron has its own weight vector, so the output of the layer is an $m$-dimensional vector computed as the input vector pre-multiplied by an $m\times n$ matrix of weights.
That is, the output is an $m$-dimensional vector that is the linear transformation of an $n$-dimensional input vector.
The output of each neuron is in effect a linear classifier where the weight vector defines a borderline between two classes and where the input vector lies some distance to one side of it or the other.
The combined result of all $m$ neurons is an $m$-dimensional hyperplane that independently classifies the $n$ dimensions of the input into two $m$-dimensional classes in the output.
If the weights are derived via least mean-squared (LMS) estimation from matched pairs of input-output data they form a linear regression, \textit{i.e.}\ the hyperplane that is closest in the LMS sense to all the outputs given the inputs.

The hyperplane maps new input points to output points that are consistent with the original data, in the sense that some \textit{error function} between the computed outputs and the actual outputs in the training data is minimized.
Multiple layers of linear maps, wherein the output of one linear classifier or regression is the input of another, is actually equivalent to a different single linear classifier or regression.
This is because the output of $k$ different layers reduces to the multiplication of the inputs by
a single $q\times n$ matrix that is the product of the $k$ matrices, one per layer.

To classify inputs non-linearly or to approximate a nonlinear function with a regression, each neuron adds a numerical bias value to the result of its input sum of products (the linear classifier) and passes that through a nonlinear \textit{activation function}.
The actual form of the activation function is a design parameter.
But they all have the characteristic that they map the real line through a monotonic increasing function
that has an inflection point at zero.
In a single neuron, the bias effectively shifts the inflection point of the activation function to the value of the bias itself.
So the sum of products is mapped through an activation function centered on the bias. 
Any pair of activation functions so defined are capable of producing a pulse between their inflection points if each one is scaled and one is subtracted from the other.
In effect each pair of neurons samples the input space and outputs a specific value for all inputs within the limits of the pulse.
Given training data consisting of input-output pairs -- input vectors each with a corresponding output vector -- the ANN learns an approximation to the function that produced each of the outputs from its corresponding input.
That approximation is the partition of the input space into samples that minimizes the error function
between the output of the ANN given its training inputs and the training outputs.
This is stated mathematically by the \textit{universal approximation theorem} which implies that
any functional mapping between input vectors and output vectors can be approximated to with arbitrary accuracy with an ANN provided that it has a sufficient number of neurons in a sufficient number of layers with a specific activation function \cite{Cybenko1989} \cite{HORNIK1991251} \cite{DBLP:journals/corr/abs-1709-02540} \cite{Hanin-2017}.

Given the dimensions of the input and output vectors, the number of layers, the number of neurons in each layer, the form of the activation function, and an error function, the weights are computed via optimization over input-output pairs to minimize the error function.
That way the resulting network is a best approximation of the known input-output data.

\subsection {\textbf{How do these networks learn?}} 

\begin{figure}
  \begin{center}
  \includegraphics[scale=0.4]{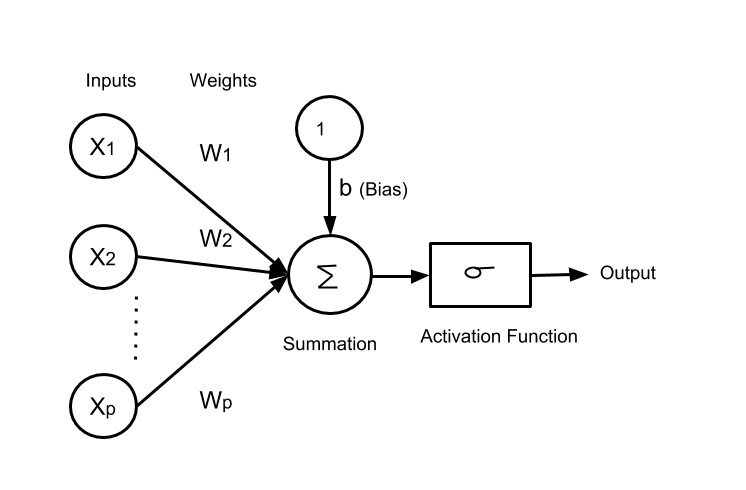}
   \caption{The Perceptron Learning Model}
  \label{fig:perceptron}
  \end{center}
\end{figure}

Neural networks are capable of learning - by changing the distribution of weights it is possible to approximate a function representative of the patterns in the input. The key idea is to re-stimulate the black-box using new excitation (data) until a sufficiently well-structured representation is achieved. Each stimulation redistributes the neural weights a little bit - hopefully in the right direction, given the learning algorithm involved is appropriate for use, until the error in approximation w.r.t some well-defined metric is below a practitioner-defined lower bound. Learning then, is the aggregation of a variable length of causal chains of neural computations \cite{SCHMIDHUBER201585} seeking to approximate a certain pattern recognition task through linear/nonlinear modulation of the activation of the neurons across the architecture. The instances in which chains of implicit linear activation fail to learn the underlying structure, non-linearity aids the modulation process. The term \textit{'deep'} in this context is a direct indicator of the space complexity of the aggregation chain across many \textit {hidden layers} to learn sufficiently detailed representations. Theorists and empiricists alike have contributed to an exponential growth in studies using {Deep} Neural Networks, although generally speaking, the existing constraints of the field are well-acknowledged \cite{marcus1} \cite{papernot1} \cite{abbe1}. Deep learning has grown to be one of the principal components of contemporary research in artificial intelligence in light of its ability to scale with input data and its capacity to generalize across problems with similar underlying feature distributions, which are in stark contrast to the hard-coded, problem-specific pattern recognition architectures of yesteryear.

\vspace{2 em}

\begin{table}[h!]
\vspace{2 em}

\caption{Some Key Advances in Neural Networks Research}
\centering
 \begin{tabular}{|m{1.5cm} | m{20em}| } 
 \hline
 \centering \textbf{People Involved} & \textbf{Contribution}\\ \hline 

\centering McCulloch \& Pitts & ANN models with adjustible weights (1943) \cite{McCulloch1943} \\
 \hline
 \centering Rosenblatt & The Perceptron Learning Algorithm (1957) \cite{Rosenblatt58theperceptron:} \\
 \hline
 \centering Widrow and Hoff & Adaline (1960), Madaline Rule I (1961) \& Madaline Rule II (1988)\cite{23872} \cite{58323}\\  
 \hline
\centering Minsky \& Papert & The XOR Problem (1969) \cite{Minsky1969PerceptronsA}\\  
\hline
\centering Werbos (Doctoral Dissertation) & Backpropagation (1974) \cite{werbos1994roots} \\ 
\hline
\centering Hopfield & Hopfield Networks (1982) \cite{Hopfield2554} \\ 
\hline
\centering Rumelhart, Hinton \& Williams & Renewed interest in backpropagation: multilayer adaptive backpropagation (1986) \cite{Rumelhart:1986:LIR:104279.104293}\\
\hline
\centering Vapnik, Cortes & Support Vector Networks (1995) \cite{Cortes1995}\\
\hline
\centering Hochreiter \& Schmidhuber & Long Short Term Memory Networks (1997) \cite{doi:10.1162/neco.1997.9.8.1735} \\
\hline

\centering LeCunn et. al. & Convolutional Neural Networks (1998) \cite{lecun1998gradient}\\
\hline

\centering Hinton \& Ruslan & Hierarchical Feature Learning in Deep Neural Networks (2006) \cite{hinton2006fast} \\

\hline
 
 \end{tabular}
 \label{tab:symbolsummary}

\end{table}


\subsection {\textbf{Why are deep neural networks garnering so much attention now?}}
Multi-layer neural networks have been around through the better part of the latter half of the previous century. A natural question to ask why deep neural networks have gained the undivided attention of academics and industrialists alike in recent years? There are many factors contributing to this meteoric rise in research funding and volume. Some of these are briefed:
\vspace{1 em}

\begin{itemize}
    \item A surge in the availability of large training data sets with high quality labels
    \vspace{1 em}
    \item Advances in parallel computing capabilities and multi-core, multi-threaded implementations
\vspace{1 em}
    \item Niche software platforms such as PyTorch \cite{paszke2017automatic}, Tensorflow \cite{tensorflow2015-whitepaper}, Caffe \cite{Jia:2014:CCA:2647868.2654889} , Chainer \cite{chainer_learningsys2015}, Keras \cite{chollet2015keras}, BigDL \cite{Dai2018BigDLAD} etc. that allow seamless integration of architectures into a GPU computing framework without the complexity of addressing low-level details such as derivatives and environment setup. Table \ref{tab:summary} provides a summary of popular Deep Learning Frameworks.
\vspace{1 em}
    \item Better regularization techniques introduced over the years help avoid overfitting as we scale up: techniques like batch normalization, dropout, data augmentation, early stopping etc are highly effective in avoiding overfitting and can improve model performance with scaling. 
\vspace{1 em}
    \item Robust optimization algorithms that produce near-optimal solutions: Algorithms with adaptive learning rates (AdaGrad, RMSProp, Adam, Adaboost), Stochastic Gradient Descent (with standard momemtum or Nesterov momentum), Particle Swarm Optimization, Differential Evolution, etc. 
\end{itemize}

\begin{table}[h]

\caption{A Collection of Popular Deployment Platforms }
\centering
 \begin{tabular}{|m{1.5 cm} | m{7 cm}| } 
 \hline
 \centering \textbf{Software Platform} & \textbf{Purpose}\\ \hline

\centering Tensorflow \cite{tensorflow2015-whitepaper} & Software library with high performance numerical computation and support for Machine Learning and Deep Learning architectures compatible to be deployed in CPU, GPU and TPU.

url: \url{https://www.tensorflow.org/}
\\ \hline

\centering Theano \cite{2016arXiv160502688short} & GPU compatible Python library with tight integration to NumPy involves smooth mathematical operations on multidimensional arrays.

url: \url{http://deeplearning.net/software/theano/}
\\ \hline

\centering CNTK \cite{Seide:2016:CMO:2939672.2945397} & Microsoft Cognitive Toolkit (CNTK) is a Deep Learning Framework describing computations through directed graphs.

url: \url{https://www.microsoft.com/en-us/cognitive-toolkit/}
\\ \hline

\centering Keras \cite{chollet2015keras} & It runs on top of Tensorflow, CNTK or Theano compatible to be deployed in CPU and GPU.

url: \url{https://keras.io/}
\\ \hline

\centering PyTorch \cite{paszke2017automatic} & 
Distributed training and performance evaluation platform integrated with Python supported by major cloud platforms.

url: \url{https://pytorch.org/}
\\ \hline

\centering Caffe \cite{Jia:2014:CCA:2647868.2654889} & Convolutional Architecture for Fast Feature Embedding (Caffe) is a Deep Learning framework with focus on image classsification and segmentation and deployable in both CPU and GPU.

url: \url{http://caffe.berkeleyvision.org/}
\\ \hline

\centering Chainer \cite{chainer_learningsys2015} &
Supports CUDA computation and multiple GPU implementation.

url: \url{https://chainer.org/}
\\ \hline

\centering BigDL \cite{Dai2018BigDLAD} & 
Distributed deep learning library for Apache Spark supporting programming languages Scala and Python.

url: \url{https://software.intel.com/en-us/articles/bigdl-distributed-deep-learning-on-apache-spark}

\\ \hline

 \end{tabular}
 \label{tab:summary}
 
\end{table}

\begin{figure*}[t]
  \begin{center}
  \includegraphics[scale=.7]{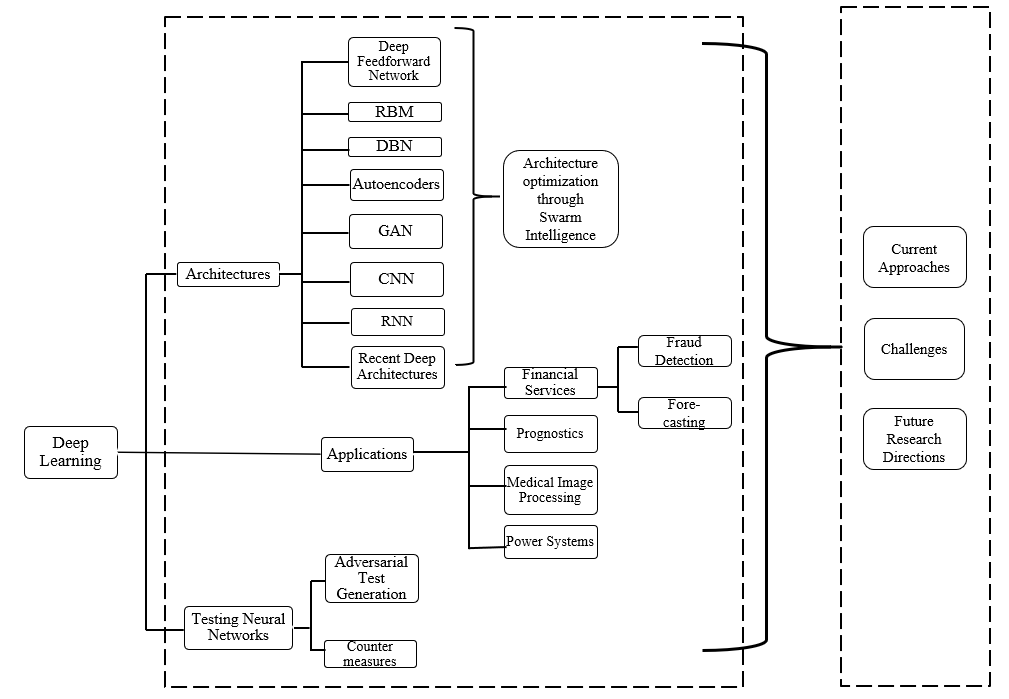}
  \caption{Organization of the Review}
\label{fig:LAYOUT}
  \end{center}
\end{figure*}


\subsection {\textbf{Review Methodology}}

The article, in its present form serves to present a collection of notable work carried out by researchers in and related to the \textit{deep learning} niche. It is by no means exhaustive and limited in its own right to capture the global scheme of proceedings in the ever-evolving complex web of interactions among the deep learning community. While cognizant of the difficulty of achieving the stated goal, we tried to present nonetheless to the reader an overview of pertinent scholarly collections in varied niches in a single article. 

\vspace{0.5 cm}
The article makes the following contributions from a practitioner's reading perspective:
\vspace{1 em}
\begin{itemize}
\item It walks through foundations of biomimicry involving artificial neural networks from biological ones, commenting on how neural network architectures learn and why deeper layers of neural units are needed for certain of pattern recognition tasks.
\vspace{1 em}
\item It talks about how several different deep architectures work, starting from Deep feed-forward networks (DFNNs) and Restricted Boltzmann Machines (RBMs) through Deep Belief Networks (DBNs) and Autoencoders. It also briefly sweeps across Convolutional neural networks (CNNs), Recurrent Neural Networks (RNNs), Generative Adversarial Networks (GANs) and some ot the more recent deep architectures. This cluster within the article serves as a baseline for further readings or as a refresher for the sections which build on it and follow.  
\vspace{1 em}
\item The article surveys two major computational areas of research in the present day deep learning community that we feel have not been adequately surveyed yet - (a) Multi-agent approaches in automatic architecture generation and learning rule optimization of deep neural networks using swarm intelligence and (b) Testing, troubleshooting and robustness analysis of deep neural architectures which are of prime importance in guaranteeing up-time and ensuring fault-tolerance in mission-critical applications.    
\vspace{1 em}

\item A general survey of developments in certain application modalities is presented. These include:

 \begin{enumerate}
 \item Anomaly Detection in Financial Services, 
 
 \item Financial Time Series Forecasting, 
 
 \item Prognostics and Health Monitoring, 
 
 \item Medical Imaging and 
 
 \item Power Systems
 \end{enumerate}
\end{itemize}
\vspace{1 em}

Figure \ref{fig:LAYOUT} captures a high-level hierarchical abstraction of the organization of the review with emphasis on current practices, challenges and future research directions. The content organization is as follows: Section \ref{Architectures} outlines some commonly used deep architectures with a high-level working mechanisms of each, Section \ref{swarmdeep} talks about the infusion of swarm intelligence techniques within the context of deep learning and Section \ref{testingneural} details diagnostic approaches in assuring fault-tolerant implementations of deep learning systems. Section \ref{applications} makes an exploratory survey of several pertinent applications highlighted in the previous paragraph while Section \ref{conclusions} makes a critical dissection of the general successes and pitfalls of the field as of now and Section \ref{sec:conclusion} concludes the article.


\section{Deep architectures: Working mechanisms}\label{Architectures}
There are numerous deep architectures available in the literature. The Comparison of architectures is difficult as different architectures have different advantages based on the application and the characteristics of the data involved, for example, In vision, Convolutional Neural Networks \cite{lecun1998gradient}, for sequences and time series modelling Recurrent neural networks \cite{kombrink2011recurrent} is prefered. However, deep learning is a fast evolving field. In every year various architectures with various learning algorithms are developed to endure the need to develop human-like efficient machines in different domains of application.


\subsection {\textbf{Deep Feed-forward Networks}}

\begin{figure}
  \begin{center}
  \includegraphics[scale=0.25]{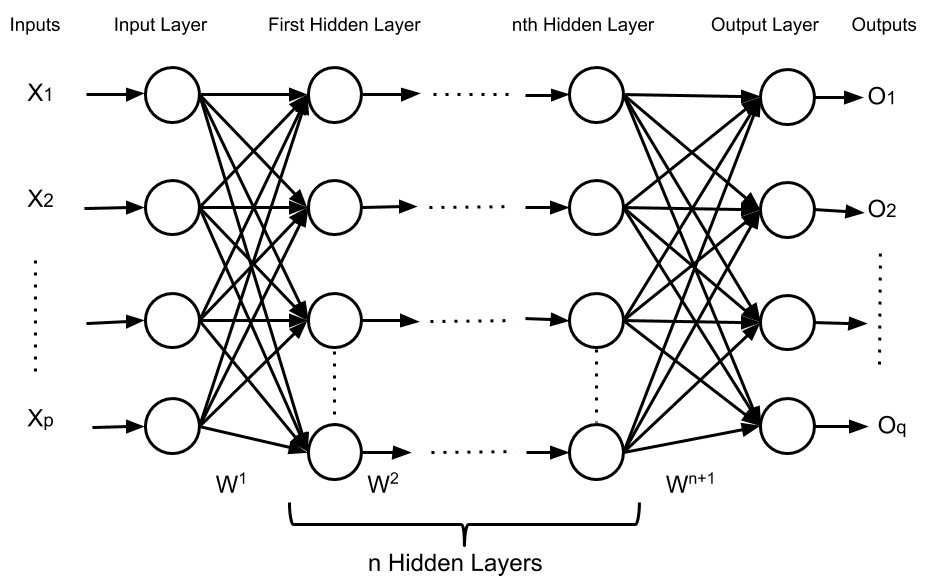}
   \caption{Deep Feed-forward Neural Network with n Hidden layers, p input units and q output units with weights W.}
  \label{fig:FFN}
  \end{center}
\end{figure}

Deep Feedforward Neural network, the most basic deep architecture with only the connections between the nodes moves forward. Basically, when a multilayer neural network contains multiple numbers of hidden layers, we call it deep neural network \cite{deng2014deep}. An example of Deep Feed-Forward Network with n hidden layers is provided in Figure \ref{fig:FFN}. Multiple hidden layers help in modelling complex nonlinear relation more efficiently compared to the shallow architecture. A complex function can be modelled with less number of computational units compared to a similarly performing shallow network due to the hierarchical learning possible with the multiple levels of nonlinearity \cite{bengio2009learning}.  Due to the simplicity of architecture and the training in this model, It is always a popular architecture among researchers and practitioners in almost all the domains of engineering. Backpropagation using gradient descent \cite{rumelhart1985learning} is the most common learning algorithm used to train this model.  The algorithm first initialises the weights randomly, and then the weights are tuned to minimise the error using gradient descent. The learning procedure involves multiple forward and backwards passes consecutively. In forward pass, we forward the input towards the output through multiple hidden layers of nonlinearity and ultimately compare the computed output with the actual output of the corresponding input. In the backward pass, the error derivatives with respect to the network parameters are back propagated to adjust the weights in order to minimise the error in the output. The process continues multiple times until we obtained a desired improvement in the model prediction. If $X_i$ is the input and $f_i$ is the nonlinear activation function in layer i, the output of the layer i can be represented by,
\begin{equation}
    X_{i+1}=f_i(W_iX_i+b_i)
\end{equation}

$X_{i+1}$, as this becomes input for the next layer. $W_i$ and $b_i$are the parameters connecting the layer i with the previous layer. In the backward pass, these parameters can be updated with,

\begin{equation}
    {W_{new}=W-\eta \partial E/\partial W}
\end{equation}

\begin{equation}
    b_{new\ \ }=b-\eta \partial E/\partial b 
\end{equation}

Where $W_{new}$ and $b_{new}$ are the updated parameters for W and b respectively, and E is the cost function and $\eta $is the learning rate. Depending on the task to be performed like regression or classification, the cost function of the model is decided. Like for regression, root mean square error is common and for classification softmax function.

Many issues like overfitting, trapped in local minima and vanishing gradient issues can arise if a deep neural network is trained naively. This was the reason; neural network was forsaken by the machine learning community in the late 1990s. However, in 2006 \cite{hinton2006fast, bengio2007greedy}, with the advent of unsupervised pretraining approach in deep neural network, the neural network is revived again to be used for the complex tasks like vision and speech. Lately, many other techniques like l1, l2 regularisation \cite{bengio2013advances}, dropout \cite{dahl2013improving}, batch normalisation \cite{ioffe2015batch}, good set of weight initialisation techniques \cite{sussillo2014random, mishkin2015all, glorot2010understanding, kumar2017weight} and good set of activation functions \cite{maas2013rectifier} are introduced to combat the issues in training deep neural networks.


\subsection {\textbf{Restricted Boltzmann Machines}}
Restricted Boltzmann Machine (RBM) \cite{fischer2012introduction} can be interpreted as a stochastic neural network. It is one of the popular deep learning frameworks due to its ability to learn the input probability distribution in supervised as well as unsupervised manner. It was first introduced by Paul Smolensky in 1986 with the name Harmonium \cite{smolensky1986information}. However, it gets popularised by Hinton in 2002 \cite{hinton2002training} with the advent of the improved training algorithm to RBM.  After that, it got a wide application in various tasks like representation learning \cite{coates2011analysis}, dimensionality reduction \cite{hinton2006reducing}, prediction problems \cite{larochelle2008classification}. However, deep belief network training using the RBM as building block \cite{hinton2006fast} was the most prominent application in the history of RBM that provides the starting of deep learning era. Recently RBM is getting immense popularity in the field of collaborative filtering \cite{salakhutdinov2007restricted} due to the state of the art performance in Netflix.

\begin{figure}
  \begin{center}
  \includegraphics[scale=0.25]{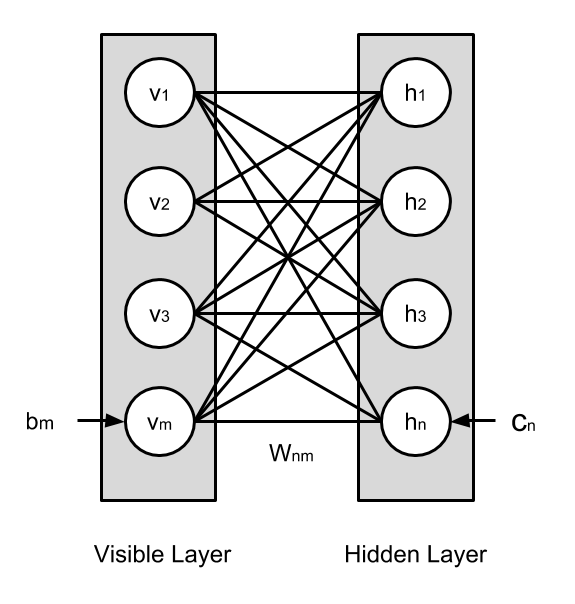}
   \caption{RBM with m visible units and n hidden units}
  \label{fig:RBM}
  \end{center}
\end{figure}

Restricted Boltzmann Machine is a variation of Boltzmann machine with the restriction in the intra-layer connection between the units, and hence called restricted. It is an undirected graphical model containing two layers, visible and hidden layer, forms a bipartite graph. Different variations of RBMs have been introduced in literature in terms of improving the learning algorithms, provided the task. Temporal RBM \cite{sutskever2007learning} and conditional RBM \cite{taylor2007modeling} proposed and applied to model multivariate time series data and to generate motion captures, Gated RBM \cite{memisevic2007unsupervised} to learn transformation between two input images, Convolutional RBM \cite{lee2009convolutional, lee2009unsupervised} to understand the time structure of the input time series, mean-covariance RBM \cite{dahl2010phone, hinton2010modeling, mohamed2012understanding} to represent the covariance structure of the data, and many more like Recurrent TRBM \cite{sutskever2009recurrent}, factored conditional RBM (fcRBM) \cite{taylor2009factored}. Different types of nodes like Bernoulli, Gaussian \cite{hinton2012practical} are introduced to cope with the characteristics of the data used. However, the basic RBM modelling concept introduced with Bernoulli units. Each node in RBM is a computational unit that processes the input it receives to make stochastic decisions whether to transmit that input or not. An RBM with m visible and n hidden units is provided in Figure \ref{fig:RBM}.

The joint probability distribution of an standard RBM can be defined with Gibbs distribution \textbf{ }$p(v,h)\ =\frac{1}{Z}e^{-E(v,h)}\ $, where energy function E(v,h)  can be represented with:

\begin{equation}
  E(v,h)\ =\ -\sum^n_{i=1}{}\sum^m_{j=1}{}w_{ij}h_jv_i\ -\sum^m_{j=1}{}b_jv_j\ -\ \sum^n_{i=1}{}c_ih_i
\end{equation}

Where, m,n are the number of visible and hidden units, $v_j$, $h_j$ are the states of the visible unit j and hidden unit i, $b_j$, $c_j$ are the real-valued biases corresponding to the jth visible unit and ith hidden unit respectively, $w_{ij}$ is real-valued weights connecting visible units with hidden units. Z is the normalisation constant (sum over all the possible combinations for $e^{-E(v,h)}$)  to ensure the probability distributions sums to 1. The restriction made in the intralayer connection make the RBM hidden layer variables independent given the states of the visible layer variables and vice versa. This easy down the complexity of modelling the probability distribution and hence the probability distribution of each variable can be represented by conditional probability distribution as given below:

\begin{equation}
    p(h|v)=\prod^n_{i=1}{}p(h_i|v)
\end{equation}    
\begin{equation}
    \space p(v|h)=\prod^m_{j=1}{}p(v_j|h)
\end{equation}

RBM is trained to maximise the expected probability of the training samples. Contrastive divergence algorithm proposed by Hinton \cite{hinton2002training} is popular for the training of RBM. The training brings the model to a stable state by minimising its energy by updating the parameters of the model. The parameters can be updated using the following equations:

\begin{equation}
    \mathit{\Delta}w_{ij}=\epsilon ({<v_ih_j>}_{data}-{<v_ih_j>}_{model})
\end{equation}

\begin{equation}
    \mathit{\Delta}b_i=\epsilon ({<v_i>}_{data}-{<v_i>}_{model})
\end{equation}

\begin{equation}
    \mathit{\Delta}c_j=\epsilon ({<h_j>}_{data}-{<h_j>}_{model})
\end{equation}

Where, $\epsilon $ is the learning rate, $<$ . $>$ data , $<$ . $>$ model are used to represent the expected values of the data and the model.


\subsection {\textbf{Deep Belief Networks}}

\begin{figure}
  \begin{center}
  \includegraphics[scale=0.25]{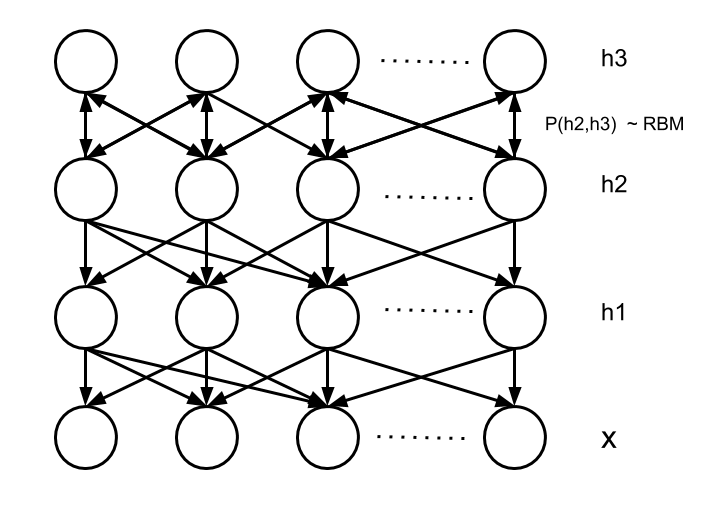}
  \caption{DBN with input vector \textbf{x} with 3 hidden layers}
  \label{fig:DBN}
  \end{center}
\end{figure}

Deep belief network (DBN) is a generative graphical model composed of multiple layers of latent variables. The latent variables are typically binary, can represent the hidden features present in the input observations. The connection between the top two layers of DBN is undirected like an RBM model, hence a DBN with 1 hidden layer is just an RBM. The other connections in DBN except last are directed graphs towards the input layer. DBN is a generative model, hence to generate a sample from DBN follows a top-down approach. We first draw samples from the RBM on the top layer, this is usually done by Gibbs sampling, then we can perform sampling from the visible units by a simple pass of ancestral sampling in a top-down fashion. A standard DBN model \cite{goodfellow2016deep} with three hidden layers is shown in Figure \ref{fig:DBN}.

Inference in DBN is an intractable problem due to the explaining away effect in the latent variable model. However, in 2006 Hinton \cite{hinton2006fast} proposed a fast and efficient way of training DBN by stacking Restricted Boltzmann Machine (RBM) one above the other. The lowest level RBM during training learns the distribution of the input data. The next level of RBM block learns high order correlation between the hidden units of the previous hidden layer by sampling the hidden units. This process repeated for each hidden layer till the top. A DBN with L numbers of hidden layer models the joint distribution between its visible layer v and the hidden layers $h^l$, where l =1,2, {\dots} L as follows:

\begin{equation}
    p(v,\ h^1,...\ ,\ h^L)\ =\ p(v|\ h^1)(\prod^{L-2}_{l=1}{}p(\ h^l|\ h^{l+1}))p(\ h^{L-1},\ h^L)
\end{equation}

The log-probability of the training data can be improved by adding layers to the network, which, in turn, increases the true representational power of the network \cite{le2008representational}. The DBN training proposed in 2006 \cite{hinton2006fast} by Hinton led to the deep learning era of today and revived the neural network. This was the first deep architecture in the history able to train efficiently. Before that, it was almost impossible to train deep architectures. Deep architectures build by initialising the weights with DBN, outperformed the kernel machines, that was in the research landscape at that time. DBN, along with its use as generative models, significantly applied as discrimination model by appending a discrimination layer at the end and fine-tuning the model using the target labels provided \cite{lecun2015deep}. In most of the applications, this approach of pretraining a deep architecture led to the state of the performance in discriminative model \cite{hinton2005kind, hinton2006fast, bengio2007greedy, poultney2007efficient, hinton2006reducing} like in recognising handwritten digits, detecting pedestrians, time series prediction etc. even when the number of labelled data was limited \cite{sermanet2013pedestrian}. It has got immense popularity in acoustic modelling \cite{mohamed2012acoustic} recetly as the model could provide upto 20\% improvement over state of the art models, Hidden Markov Model, Gaussian Mixture Model. The approach creates feature detectors hierarchically as “features of features” in pretraining that provide a good set of initialised weights to the discriminative model. The initialised weights are in a region near the optimal weights that can improve both modelling and the convergence in fine-tuning \cite{hinton2005kind, erhan2010does}. DBN has been used as an initialised model in classification in many applications like in phone recognition \cite{dahl2010phone}, computer vision \cite{hinton2010modeling} where it is used for the training of higher order factorized Boltzmann machine, speech recognition \cite{siniscalchi2013hermitian, siniscalchi2013exploiting, yu2012boosting} for pretraining DNN, for pretraining of deep convolutional neural network (CNN) \cite{lee2009convolutional, lee2011unsupervised, lee2009unsupervised}. The improved performance is due to the ability to learn some abstract features by the hidden layer of the network. Some of the work on analysis of the features to understand what is lost and what is captured during its training is demonstrated in \cite{mohamed2012understanding, susskind2011deep, stoyanov2011empirical}.


\subsection {\textbf{Autoencoders}}
\begin{figure}
  \begin{center}
  \includegraphics[scale=0.25]{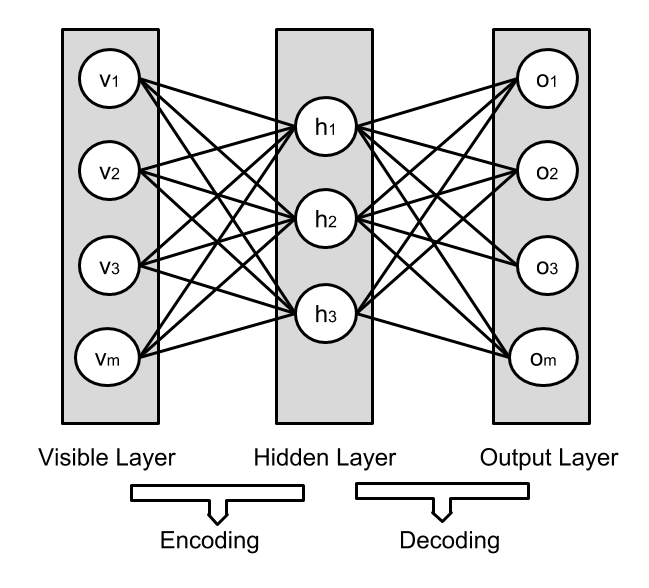}
  \caption{Autoencoder with 3 neurons in hidden layer}
  \label{fig:AE}
  \end{center}
\end{figure}

Autoencoder is a three-layer neural network, as shown in Figure \ref{fig:AE}, that tries to reconstruct its input in its output layer. Hence, the output layer of an autoencoder contains the same number of units as the input layer. The hidden layer typically contains less number of neurons compared to the visible layer, tries to encode or represent the input in a more compact form. It shares the same idea as RBM, but it typically uses deterministic distribution instead of stochastic units with particular distribution as in the case of RBM.

Like feedforward neural network, autoencoder is typically trained using backpropagation algorithm. The training consists of two phases: Encoding and Decoding. In the encoding phase, the model tries to encode the input into some hidden representation using the weight metrics of the lower half layer, and in the decoding phase, it tries to reconstruct the same input from the encoding representation using the metrics of the upper half layer. Hence, weights in encoding and decoding are forced to be the transposed of each other. The encoding and decoding operation of an autoencoder can be represented by equations below:
In encoding phase,
\begin{equation}
    y'=f(wx+b)
\end{equation}

Where w, b are the parameters to be tuned, f is the activation function, x is the input vector, and y is the hidden representation.
In decoding phase,
\begin{equation}
    x'=f(w'y'+c)
\end{equation}

Where$\ w'$ is the transpose of $w,c$ is the bias to the output layer, $x'$ is the reconstructed input at the output layer. The parameters of the autoencoder can be updated using the following equations:

\begin{equation}
    {w_{new}=w-\eta \partial E/\partial w}
\end{equation}

\begin{equation}
    b_{new\ \ }=b-\eta \partial E/\partial b
\end{equation}

Where $w_{new}$ and $b_{new}$ are the updated parameters for w and b respectively at the end of the current iteration, and E is the reconstruction error of the input at the output layer.

Autoencoder with multiple hidden layers forms a deep autoencoder. Similar like in deep neural network, autoencoder training may be difficult due to multiple layers. This can be overcome by training each layer of deep autoencoder as a simple autoencoder \cite{hinton2006fast, bengio2007greedy}. The approach has been successfully applied to encode documents for faster subsequent retrieval \cite{salakhutdinov2009semantic}, image retrieval, efficient speech features \cite{deng2010binary} etc. As like RBM stacking to form DBN \cite{hinton2006fast} for layerwise pretraining of DNN, autoencoder \cite{bengio2007greedy} along with sparse encoding energy-based model \cite{poultney2007efficient} are independently developed at that time. They both were effectively used to pre-train a deep neural network, much like the DBN. The unsupervised pretraining using autoencoder has been successfully applied in many fields like in image recognition and dimensionality reduction in MNIST \cite{hinton2006reducing, deng2010binary, deng2012mnist}, multimodal learning in speech and video images \cite{ngiam2011learning, ngiam2011multimodal} and many more. Autoencoder has got immense popularity as generative model in recent years \cite{deng2014deep, kingma2013auto}. Non Probabilistic and non-generative nature of conventional autoencoder has been generalised to generative modelling \cite{alain2014regularized, bengio2013advances, bengio2013representation, bengio2014deep, bengio2013deep} that can be used to generate the samples from the network meaningfully.

Several variations of autoencoders are introduced with quite different properties and implementation to learn more efficient representation of data. One of the popular variation of autoencoder that is robust to input variations is denoising autoencoder \cite{bengio2014deep, bengio2013advances, bengio2013deep}. The model can be used for good compact representation of input with the number of hidden layers less than the input layer. It can also be used to perform robust modelling of the input distribution with higher number of neurons in the hidden layer. The robustness in denoising autoencoder is achieved by introducing dropout trick or by introducing some gaussian noise to the input data \cite{vincent2011connection, vincent2010stacked} or to the hidden layers \cite{hinton2012improving}. The approach helps in many many ways to improve performance. It virtually increasing the training set hence reduce overfitting, and make robust representation of the input. Sparse autoencoder \cite{hinton2012improving} is introduced in a consideration to allow larger number of hidden units than the visible units to make it easier and efficient representation of the input distribution in the hidden layer. The larger hidden layer represent the input representation by turning on and off the units in the hidden layer. Variational autoencoder \cite{kingma2013auto, doersch2016tutorial} that uses quite the similar concept as RBM, learn stochastic distribution of latent variables instead of deterministic distribution. Transforming autoencoders \cite{hinton2011better} proposed as a autoencoder with transformation invariant property. The encoded features of the autoencoder can effectively reflect the transformation invariant property. The encoder is applied in image recognition \cite{hinton2011better, hinton2011transforming} purpose that contains “capsule” as the building block. “Capsule” is an independent sub-network that extracts local features within a limited window of viewing to understand if a feature entity is present with certain probability. Pretraining for CNN using regularised deep autoencoder is very much popularised in recent years in computer vision works.  Robust models of CNN is obtained with denoising autoencoder \cite{bengio2013representation} and sparse autoencoder with pooling and local contrast normalization \cite{le2013building} which provides not only translation-invariant features but also scaling and out-of-plane rotation invariant features.


\begin{figure*}[t]
  \begin{center}
  \includegraphics[scale=.35]{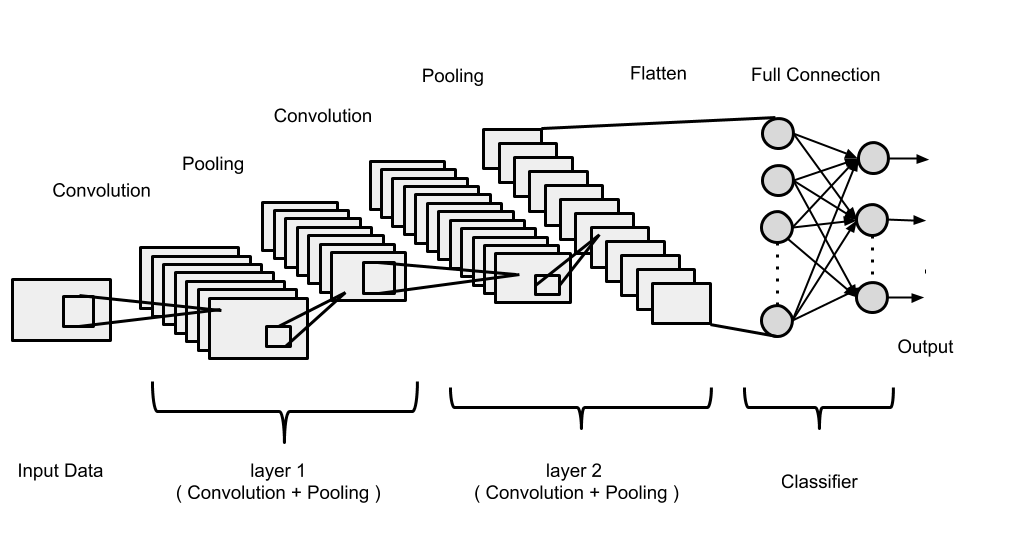}
  \caption{Convolution and Pooling Layers in a CNN}
\label{fig:CNN}
  \end{center}
\end{figure*}

\begin{figure}[t]
\label{fig:2}
  \begin{center}
   \includegraphics[scale=0.4]{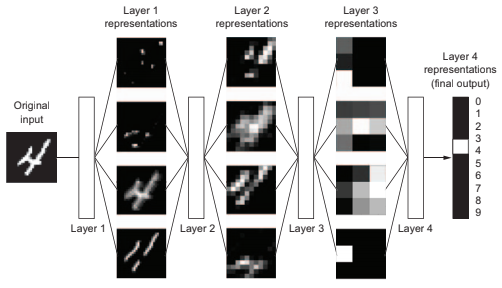}
  \caption{Representations of an image of handwritten digit learned by CNN}\label{waveforms}
  \end{center}
\end{figure}


\subsection {\textbf{Convolutional Neural Networks}} 
Convolutional Neural Networks are a class of neural networks that are extremely good for processing images. Although its idea was proposed way back in 1998 by LeCun et. al in their paper entitled "Gradient-based learning applied to document recognition" \cite{726791} but the deep learning world actually saw it in action when Krizhevsky et. al were able win the ILSVRC-2012 competition. The architecture that Krizhevsky et. al proposed is popularly known as AlexNet \cite{Krizhevsky:2012:ICD:2999134.2999257}. This remarkable win started the new era of artificial intelligence and the computation community witnessed the real power of CNNs. Soon after this, several architectures have been proposed and still are being proposed. And in many cases, these CNN architectures have been able to beat human recognition power as well. It is worth to note that, The deep learning revolution actually with the usage of Convolutional Neural Networks (CNNs). 
CNNs are are extremely useful for a set computer vision related tasks such as image detection, image segmentation, image classification and so on and all of these tasks are practically well aligned. 
On a very high level, deep learning is all about learning data representations and in order to do so deep learning systems typically breaks down complex representations into a set of simpler representations. As mentioned earlier, CNNs are particularly useful when it comes to images as images have a special spatial property in their formations. An image has several characteristics like edges, contours, strokes, textures, gradients, orientation, colour. A CNN breaks down an image in terms of simple properties like these and learn them as representations in different layers \cite{DBLP:journals/nature/LeCunBH15}. Figure \ref{waveforms} is a good representative of this learning scheme. 

The layers involved in any CNN model are the convolution layers and the subsampling/pooling layers which allow the network learn filters that are specific to specific parts in an image. The convolution layers help the network retain the spatial arrangement of pixels that is present in any image whereas the pooling layers allow the network to summarize the pixel information \cite{Goodfellow_Book}. There are several CNN architectures ZFNet, AlexNet, VGG, YOLO, SqueezeNet, ResNet and so on and some these have been discussed in section \ref{DeepArch}. 

\subsection {\textbf{Recurrent Neural Networks}}

\begin{figure*}[t]
\label{fig:RNN}
  \begin{center}
  \includegraphics[scale=.35]{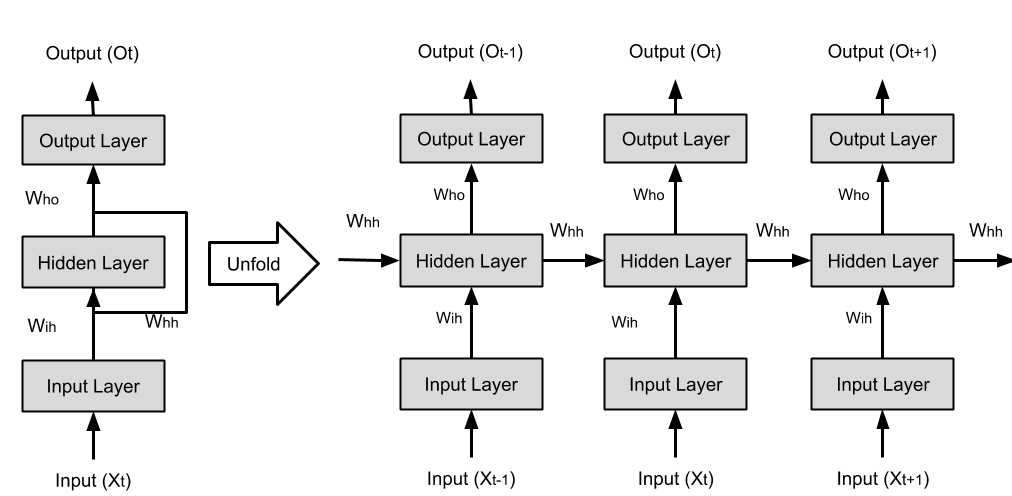}
  \caption{A Recurrent Neural Network Architecture}
  \end{center}
\end{figure*}

Although Hidden Markov Models (HMM) can express time dependencies, they become computationally unfeasible in the process of modelling long term dependencies which RNNs are capable of. A detailed derivation of Recurrent Neural Network from differential equations can be found in \cite{Sherstinsky2018FundamentalsOR}. RNNs are form of feed-forward networks spanning adjacent time steps such that at any time instant a node of the network takes the current data input as well as the hidden node values capturing information of previous time steps. During the backpropagation of errors across multiple timesteps the problem of vanishing and exploding gradients take place which can be avoided by Long Short Term Memory (LSTM) Networks introduced by Hochreiter and Schmidhuber \cite{Hochreiter}. The amount of information to be retained from previous time steps is controlled by a sigmoid layer known as `forget' gate whereas the sigmoid activated `input gate' decides upon the new information to be stored in the cell followed by a hyperbolic tangent activated layer to produce new candidate values which is updated taking forget gate coefficient weighted old state's candidate value. Finally the output is produced controlled by output gate and hyperbolic tangent activated candidate value of the state.

LSTM networks with peephole connections \cite{861302} updates the three gates using the cell state information.  A single update gate instead of forget and input gate is introduced in Gated Recurrent Unit (GRU) \cite{Chung2014EmpiricalEO} merging the hidden and the cell state. In \cite{Sak2014LongSM} Sak et al., came up with training LSTM RNNs in a distributed way on multicore CPU using asynchronus SGD (Stochastic Gradient Descent) optimization for the purpose of acoustic modelling. They presented a two-layer deep LSTM architecture with each layer having a linear recurrent projection layer with more efficient use of the model parameters.
Doetch et al., \cite{Doetsch2014FastAR} proposed a LSTM based training framework composed of sequence chunks forming mini batches for training  for the purpose of handwriting recognition. With reduction of runtime by a factor of 3 the architecture uses modified gating units with layer specific weights for each gate. Palangi et al.,  \cite{Palangi2016DeepSE} implemented sentence embedding model using LSTM-RNN that sequentially extracts information from each word and embeds in a semantic vector till the end of the sentence to obtain overall semantic representation of the entire sentence. The model with capability of attenuating unimportant words and identifying salient keywords is specifically useful in web document retrieval applications.


\subsection {\textbf{Generative Adversarial Networks}}

\begin{figure}[b]
\label{fig:GAN}
  \begin{center}
  \includegraphics[scale=.35]{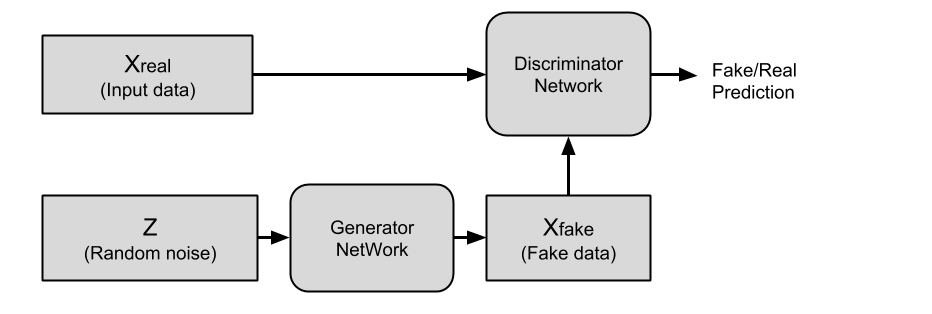}
  \caption{A Generative Adversarial Network Architecture}
  \end{center}
\end{figure}

Goodfellow et al., \cite{Goodfellow2014GenerativeAN} introduced a novel framework for Generative Adversarial Nets with simultaneous training of a generative and a discriminative model. The proposed new Generative model bypasses the difficulty of approximation of unmanageable probabilistic measures in Maximum Likelihood Estimation faced previously. The generative model tries to capture the data distribution whereas the discriminative model learns to estimate the probability of a sample either coming from training data or the distribution captured by generative model. If the two above models described by multilayer perceptrons, only backpropagation and dropout algorithms are required to train them. 

The goal in this process is to train the Generative network in a way to maximize the probability of the discriminative network to make a mistake. A unique solution can be obtained in the function space where the generative model recovers the distribution of training data and the discriminative model results into 50\% probalilty for each sample. This can be viewed as a minmax two player game between these two models as the generative models produce adversarial examples while discriminative model trying to identify them correctly and both try to improve their efficiency until the adversarial examples are indistinguishable from the original ones.

In \cite{Salimans:2016:ITT:3157096.3157346}, the authors presented training procedures to be applied to GANs focusing on producing visually sensible images. The proposed model was successful in producing MNIST samples visually indistinguishable 
from the original data and also in learning recognizable features from Imagenet dataset in a semi-supervised way. This work provides insight about appropriate evaluation metric for generative models in GANs and stable semi-supervised training approach. In \cite{Gro2017GeneralizingGA}, the authors identified distinct features of GANs from a Turing perspective. The discriminators were allowed to behave as interrogators such as in Turing Test by interacting with data sample generating processes and affirmed the increase in accuracy of the models by verification with two case studies. The first one was about inferring an agent's behavior based on a hidden stochastic process while managing its environment. The second examples talks about active self-discovery exercised by a robot to conclude about its own sensors by controlled movements.

Wu et al., \cite{Wu2016LearningAP} proposed a 3D Generative Adversarial Network (3DGAN) for three dimensional object generation using volumetric convolutional networks with a mapping from probabilistic space of lower dimension to three dimensional object space so that the 3D object can be sampled or explored without any reference image. As a result high quality 3D objects can be generated employing efficient shape descriptor learnt in an unsupervised manner by the adversarial discriminator. Vondrick et al., \cite{Vondrick2016GeneratingVW} came up with video recognition/classification and video generation/prediction model using Generative Adversarial Network (GAN) with separation of foreground from background employing spatio-temporal convolutional architecture. The proposed model is efficient in predicting futuristic versions of static images extracting meaningful features and recognizing actions embedded in the video in a minimally supervised way. Thus, learning scene dynamics from unlabeled videos using adversarial learning is the main objective of the proposed framework.

Another interesting application is generating images from detailed visual descriptions  \cite{Reed2016GenerativeAT}.
 The authors trained a deep convolutional generative
adversarial network (DC-GAN) based on encoded text features  through  hybrid character-level convolutional recurrent neural network and used manifold interpolation regularizer. The generalizability of the approach was tested by generating images from various objects and changing backgrounds.


\subsection {\textbf{Recent Deep Architectures}}
\label{DeepArch}
When it comes to deep learning and computer vision, datasets like Cats and Dogs, ImageNet, CIFAR-10, MNIST are used for benchmarking purposes. Throughout this section, the ImageNet dataset is used for the purpose of benchmarking results as it is more generalized than the other datasets just mentioned. Every year a competition named ILSVRC (ImageNet Large Scale Visual Recognition Competition) is organized (which is an image classification competition) which based on the ImageNet dataset and it is widely accepted by the deep learning community \cite{ILSVRC15}.
\\ Several deep neural network architectures have been proposed in the literature and still are being proposed with an objective of achieving general artificial intelligence. LeNet architecture, for example was proposed by Lecun et. al in 1998s  and it was originally proposed as a digit classification model. Later, LeNet has been incorporated to identify handwritten numbers on cheques \cite{726791}. Several architectures have been proposed after LeNet among which AlexNet certainly deserves to be the most notable mentions. It was proposed by Krizhevsky et. al in 2012 and AlexNet was able to beat all the competitors of the ILSVRC challenge. The discovery of AlexNet marks a significant turn in the history of deep learning for several reasons such as AlexNet incorporated the dropout regularization which was just developed by that time,  AlexNet made use of efficient GPU computing for reducing the training time which was first of its kind back in 2012 \cite{Krizhevsky:2012:ICD:2999134.2999257}. Soon after AlexNet , ZFNet was proposed by Zeiler et. al in the year of 2013 and showed state-of-the-art results on the ILSVRC challenge. It was an enhancement of the AlexNet architecture. It uses expanded mid convolution layers and incorporates smaller strides and filters in the first convolution layer for capturing the pixel information in a great detail \cite{DBLP:journals/corr/ZeilerF13}. In 2014, Google researchers came with a better model which is known as GoogleNet or the Inception Network and won the ILSVRC 2014 challenge. The main catch of this architecture is the inception layer which allows convolving in parallel with different kernel sizes. This is turn allows to learn the smaller pixel information of an image in a better way \cite{DBLP:journals/corr/SzegedyLJSRAEVR14}. It's worth to mention the VGGNet (also called VGG) architecture here. It was the runners' up in the ILSVRC 2014 challenge and was proposed by Simonyan et. al. VGG uses a 3X3 kernel throughout its entire architecture and ahieves tremendous generalization with this fixation \cite{DBLP:journals/corr/SimonyanZ14a}. The inner of the ILSVRC 2015 challenge was the ResNet architecture and was proposed by He et. al. This architecture is more formally known as Residual Networks and is deeper than the VGG architecture while still being less complex in the VGG architecture. ResNet was able to beat human performance on the ImageNet dataset and it is still being quite actively used in production \cite{DBLP:journals/corr/HeZRS15} \cite{DBLP:journals/corr/abs-1803-09820}.


\section {Swarm Intelligence in Deep Learning} \label{swarmdeep}

The introduction of heuristic and meta-heuristic algorithms in designing complex neural network architectures aimed towards tuning the network parameters to optimize the learning process has brought improvements in the performance of several Deep Learning Frameworks. In order to design the Artificial Neural Networks (ANN) automatically with evolutionary computation a Deep Evolutionary Network Structured Representation (DENSER) was proposed in \cite{Assuno2018DENSERDE}, where the optimal design for the network is achieved by a bi-leveled representation. The outer level deals with the number of layers and their sequence whereas the inner layer optimizes the parameters and hyper parameters associated with each layer defined by a context-free human perceivable grammar. Through automatic design of CNNs the proposed approach performed well on CIFER-10, CIFER-100, MNIST and Fashion MNIST dataset. On the other hand, Garro et al., \cite{Garro2015DesigningAN} proposed a methodology to automatically design ANN using basic Particle Swarm Optimization (PSO), Second Generation of Particle Swarm Optimization (SGPSO), and a New Model of PSO (NMPSO) to evolve and optimize the synaptic weights, transfer function for each neuron and the architecture itself simultaneously. The ANNs designed in this way, were evaluated over eight fitness functions. It aimed towards dimensionality reduction of the input pattern, and was compared to the traditional design architectures using well known Back-Propagation
and Levenberg-Marquardt algorithms. Das et al. \cite{DAS20143491}, used PSO to optimize the number of layers, neurons, the kind of transfer functions to be involved and the topology of ANN aimed at building channel equalizers that perform better in presence of all noise scenarios.

Wang et al. \cite{Wang2018EvolvingDC}, used  Variable-length Particle Swarm Optimization for automatic evolution of deep Convolutional Neural Network Architectures for image classification purposes. They proposed novel encoding strategy to encode CNN layers in particle vectors and introduced a Disabled layer hiding certain dimensions of the particle vector to have variable-length particles. In addition to this, to speed up the process the authors randomly picked up partial datasets for evaluation. Thus several variants of PSO along with its hybridised versions \cite{make1010010} as well as a host of recent swarm intelligence algorithms such as Quantum Double Delta Swarm Algorithm (QDDS) \cite{8628792} and its chaotic implementation \cite{jsan8010009} proposed by Sengupta et al. can be used, among others for automatic generation of architectures used in Deep Learning applications.

The problem of changing dimensionality of perceived information by each agent in the domain of Deep reinforcement learning (RL) for swarm systems has been solved in \cite{Httenrauch2018DeepRL} using an end–to–end learned mean feature embedding as state information. The research concluded that an end–to–end embedding using neural network features helps to scale up the RL architecture
with increasing numbers of agents towards better performing policies as well as ensures fast convergence.


\section {Testing neural networks\label{testingneural}}

Software employed in safety critical systems need to be rigorously tested through white-box or black-box testing. In white box testing, the internal structure of the software/program is known and utilized in generating test cases as per the test criteria/requirement. Whereas in black box testing the inputs and outputs of the program are compared as the internal code of the software cannot be accessed. Some of the previous works dealing with generating test cases revealing faulty cases can be found in \cite{529902} and in \cite{537016} using Principle component analysis. In \cite{Vanmali2002UsingAN} the authors implemented a black-box testing methodology by feeding randomly generated input test cases to an original version of a real-world test program producing the corresponding outputs, so as the input-output pairs are generated to train a neural network. Then each test case is applied to mutated and faulty version of the test program and compared against the output of the trained ANN to calculate the distance between two outputs indicating whether the faulty program has produced valid or invalid result. Thus ANN is treated as an automated ‘oracle’ which produces satisfactory results when the training set is comprised of data ensuring good coverage on the whole range of input.

 Y. Sun et al, \cite{Sun2018TestingDN} proposed a set of four test coverage criteria drawing inspiration from traditional Modified Condition/Decision Coverage (MC/DC) criteria. They also proposed algorithms for generating test cases for each criterion built upon linear programming. A new test case (an input to Deep Neural Network) is produced by perturbing a given one, where the stated algorithms should encode the test requirement and a fragment of the DNN by fixing the activation pattern obtained from the given input example, and then minimize the difference between the new and the current inputs. The utility of this method lies in bug finding, determining DNN safety statistics, measuring testing accuracy and analysis of DNN internal structure. The paper discusses about sign change, value change and distance change of a neuron pair with two neurons in adjacent layers in the context of their change in activation values in two given test cases. Four covering methods: sign sign cover, distance sign cover, sign value cover and distance value cover are explained along with test requirement and test criteria which computes the percentage of the neuron pairs that are covered by test cases with respect to the covering method.

For each test requirement an automatic test case generation algorithm is implemented based on Linear Programming (LP). The objective is to find a test input variable, whose value is to be synthesized with LP, with identical activation pattern as a given input. Hence a pair of inputs that satisfy the closeness definition are called adversarial examples if only one of them is correctly labeled by the DNN. The testing criteria necessitates that (sign or distance) changes of the condition neurons should support the (sign or value) change of every decision neuron. For a pair of neurons with a specified testing criterion, two activation patterns need to be found such that the two patterns together shall exhibit the changes required by the corresponding testing criterion. In the final test suite the inputs matching these patterns will be added.
The authors put forward results on 10 DNNs with the Sign-Sign, Distance-Sign, Sign-value and Distance-Value covering methods that show that the test generation algorithms are effective, as they reach high coverage for all covering criteria. Also, the covering methods designed are useful. This is supported by the fact that a significant portion of adversarial examples have been identified. To evaluate the quality of obtained adversarial examples, a distance curve to see how close the adversarial example is to the correct input has been plotted. It is observed that when going deeper into the DNN, it can become harder for the cover of neuron pairs. Under such circumstances, to improve the coverage performance, the use of larger data set when generating test pairs is needed. Interestingly, it seems that most adversarial examples can be found around the middle layers of all DNNs tested. In future the authors propose to find more efficient test case generation algorithms that do not require linear programming.

Katz et al. \cite{Katz2017TowardsPT}, provided methods for verifying adversarial robustness of neural networks with Reluplex algorithm, to prove, that a small perturbation to a rightly classified input should not result into misclassification. Huang et al, \cite{Huang2017SafetyVO}, proposed an
automated verification framework based on Satisfiability Modulo Theory (SMT) to test the safety of neural network by searching adversarial manipulations through exploration in the space around a given data point. The adversarial examples discovered were used to fine-tune the network.

\subsection {\textbf{Different Methods of Adversarial Test Generation}}

Despite the success of deep learning in various domains, the robustness of the architectures need to be studied before applying neural network architectures in safety critical systems. In this subsection we discuss the kind of malicious attack that can fool or mislead NN to output wrong decisions and ways to overcome them. The work presented by Tuncali et al., \cite{Tuncali2018SimulationbasedAT} deals with generating scenarios leading to unexpected behaviors by introducing perturbations in the testing conditions. For identifying fasification and critical systems behavior for autonomous driving systems, the authors focused on finding glancing counterexamples which refer to the borderline behavior of the system where it is in the verge of failing. They introduced Signal Temporal Logic (STL) formula for the problem in hand which in this case is a combination of predicates over the speed of the target car and distances of all other objects (including cars and pedestrians) and relative positions of them. Then a list of test cases is created and evaluated against STL specification. A covering array spanning all possible combinations of the values the variables can take is generated. To find a glancing behavior, the discrete parameters from the covering array that correspond to the trace that minimize STL conditions for a trace, are used to create test cases either uniformly randomly or by a cost function to guide a search over the continuous variables. Thus, a glancing test case for a trace is obtained. The proposed closed loop architecture behaves in an integrated way along with the controller and Deep Neural Network (DNN) based perception system to search for critical behavior of the vehicle. 

In \cite{Yuan2017AdversarialEA} Yuan et al discuss adversarial falsification problem explaining false positive and false negative attacks, white box attacks where there is complete knowledge about the trained NN model and black box attack where no information of the model can be accessed. With respect to adversarial specificity there are targeted and non-targeted attacks where the class output of the adversarial input is predefined in the first case and arbitrary in the second case. They also discuss about perturbation scope where individual attacks are geared towards generating unique perturbations per input whereas universal attacks generate similar attack for the whole dataset. The perturbation measurement is computed as p-norm distance between actual and adversarial input. The paper discusses various attack methods including L-BFGS attack, Fast Gradient Sign Method (FGSM) by performing update of one step gradient along the direction of the sign of the gradient of every pixel expressed as \cite{Goodfellow2014ExplainingAH}:
\begin{equation}
\eta = \epsilon sign(\nabla_x J_\theta (x,l))
\end{equation}
where $\epsilon$ is the magnitude of perturbation $\eta$ which when added to an input data generates an adversarial data.

FGSM has been extended by Basic Iterative Method (BIM) and Iterative Least-Likely Class Method (ILLC). Moosavi-Dezfooli et al. \cite{MoosaviDezfooli2016DeepFoolAS} proposed Deepfool where iterative attack was performed with linear approximation to surpass the nonlinearity in multidimensional cases.

\subsection{\textbf{Countermeasures for Adversarial Examples}}

The paper \cite{Yuan2017AdversarialEA} deals with reactive countermeasures such as Adversarial Detecting, Input Reconstruction, and Network Verification and proactive countermeasures such as Network Distillation, Adversarial (Re)training, and Classifier Robustifying. In Network Distillation high temperature softmax activation reduces the sensitivity of the model towards small perturbations. In Adversarial (Re)training adversarial examples are used during training. Adversarial detecting deals with finding the probability of a given input being adversarial or not. In input reconstruction technique a denoising autoencoder is used to transform the adversarial examples to actual data before passing them as input to the prediction module by deep NN. Also, Gaussian Process Hybrid Deep Neural Networks (GPDNNs) are proven to be more robust towards adversarial inputs. 

There are also ensembling defense strategies to counter multifaceted adversarial examples. But the defense strategies discussed here are mostly applicable to computer vision tasks, whereas the need of the day is to generate real time adversarial input detection and take measures for safety critical systems.

In \cite{Rouhani:2018:DOA:3240765.3240791} Rouhani et al., proposed an online defense framework DeepFense against adversarial deep learning. They formulated it as an unsupervised optimization problem by minimizing the less observed spaces in the latent feature hyperspace spanned by a Deep Learning network and was able to decrease the risk of integrated attacks. With integrated design of algorithms for software and hardware the proposed framework aims to maximize model reliability.

It is necessary to build robust countermeasures to be used for different types of adversarial scenarios to provide a reliable infrastructure as none of the countermeasures can be universally applicable to all sorts of adversaries. A detailed list of specific attack generation and corresponding countermeasures can be found in \cite{Chakraborty2018AdversarialAA}.

\vspace{3 em}
\begin{table}[ht!]
\caption{Distribution of Articles by Application Areas}
  \centering
 \begin{tabular}{|m{1.5cm} | m{7cm}| } 
 \hline
 \centering \textbf{Application Area} & \textbf{Authors}  \\ \hline 

 \centering Fraud Detection in Financial Services & 
Pumsirirat et al. \cite{Pumsirirat2018}, Schreyer et al. \cite{DBLP:journals/corr/abs-1709-05254}, Wang et al. \cite{Wang2018LeveragingDL}, Zheng et al. \cite{ZHENG201878}, Dong et al. \cite{DONG2018}, Gomez et al. \cite{GOMEZ2018175}, Rymantubb et al. \cite{RYMANTUBB2018130}, Fiore et al. \cite{FIORE2019448}
\\
 \hline
\centering Financial Time Series Forecasting
&  
Cavalcante et al. \cite{B1_cavalcante2016computational}, Li et al. \cite{LI20097818}, Fama et al. \cite{fred12_fama1995random}, Lu et al. \cite{LU2009115}, Tk \& Verner \cite{TKAC2016788}, Pandey et al. \cite{PANDEY2018}, Lasfer et al. \cite{lasfer2013neural}, Gudelek et al. \cite{gudelek2017deep}, Fischer \& Krauss \cite{fischer2018deep}, Santos Pinheiro \& Dras \cite{dos2017stock}, Bao et al. \cite{bao2017deep}, Hossain et al. \cite{1077}, Calvez and Cliff \cite{Calvez2018DeepLC}
\\
 \hline
\centering Prognostics and Health Monitoring
&
Basak et al. \cite{2018arXiv181008985B}, Tamilselvan \& Wang \cite{TAMILSELVAN2013124}, Kuremoto et al. \cite{KUREMOTO201447}, Qiu et al. \cite{QIU20151710}, Gugulothu et al. \cite{Gugulothu2017PredictingRU}, Filonov et al. \cite{filonov}, Botezatu et al. \cite{Botezatu:2016:PDR:2939672.2939699}
\\
 \hline
\centering Medical Image Processing
&
Suk, Lee \& Shen \cite{Suk2013LatentFR}, van Tulder \& de Bruijne \cite{7401039}, Brosch \& Tam \cite{10.1007/978-3-642-40763-5_78}, Esteva et al. \cite{esteva2017dermatologistlevel}, Rajaraman et. al. \cite{Rajaraman2018}, Kang et al. \cite{Kang20173DMC}, Hwang \& Kim \cite{HwangAndKim}, Andermatt et al. \cite{10.1007/978-3-319-46976-8_15}, Cheng et al. \cite{Cheng2018DeepSL}, Miao et al. \cite{7393571}, Oktay et al. \cite{Oktay2016MultiinputCI}, Golkov et al. \cite{7448418}
\\
 \hline
\centering Power Systems
&
Vankayala \& Rao \cite{vankayala1993artificial}, 
Chow et al. \cite{chow1988incipient}, Guo et al. \cite{guo2018deep}, Bourguet \& Antsaklis \cite{bourguet1994artificial}, Bunn \& Farmer \cite{bunn1985comparative}, Hippert et al. \cite{hippert2001neural}, Kuster et al. \cite{kuster2017electrical}, Aggarwal \& Song \cite{aggarwal1997artificial}, Zhai \cite{zhai2005time}, 
Park et al. \cite{park1991electric}, Mocanu et al. \cite{mocanu2016deep}, Chen et al. \cite{chen2018short}, 
Bouktif et al. \cite{bouktif2018optimal}, Dedinec et al.
\cite{dedinec2016deep}, Rahman et al. \cite{rahman2018predicting}, Kong et al. \cite{kong2017short}, Dong et al. \cite{dong2017short},   Kalogirou et al. \cite{kalogirou2001artificial}, Wang et al. \cite{wang2017deterministic}, Das et al. \cite{das2018forecasting}, Dabra et al. \cite{dabra2017optimization}, Liu et al. \cite{liu2015improved}, Jang et al. \cite{jang2016solar}, Gensler et al. \cite{gensler2016deep}, Abdel-Nasser et al. \cite{abdel2017accurate}, Manwell et al. \cite{manwell2010wind}, Marug\'an et al. \cite{marugan2018survey}, Wu et al. \cite{wu2016probabilistic}, Wang et al. \cite{wang2017deep}, Wang et al. \cite{wang2018deep},
Feng et al. \cite{feng2017data}, Qureshi et al.
 \cite{qureshi2017wind}
\\ \hline

\end{tabular}
\end{table}
%


\section {Applications\label{applications}}
\vspace{1 em}
\subsection {\textbf{Fraud Detection in Financial Services}}  
Fraud detection is an interesting problem in that it can be formulated in an unsupervised, a supervised and a one-class classification setting. In unsupervised learning category, class labels either unknown or are assumed to be unknown and clustering techniques are employed to figure out (i) distinct clusters containing fraudulent samples or (ii) far off fraudulent samples that do not belong to any cluster, where all clusters contained genuine samples, in which case, it is treated as an outlier detection problem. In supervised learning category, class labels are known and a binary classifier is built in order to classify fraudulent samples. Examples of these techniques include logistic regression, Naive Bayes, supervised neural networks, decision tree, support vector machine, fuzzy rule based classifier, rough set based classifier etc.  Finally, in the one-class classification category, only samples of genuine class available or fraud samples  are not considered for training even if available. These are called one-class classifiers. Examples include one-class support vector machine (aka Support vector data description or SVDD), auto association neural networks (aka auto encoders). In this category, models are trained on the genuine class data and are tested on the fraud class. Literature abounds with many studies involving traditional neural networks with various architectures to deal with the above mentioned three categories. Having said that fraud (including cyber fraud) detection is increasingly becoming menacing and fraudsters always appear to be few notches ahead of organizations in terms of finding new loopholes in the system and circumventing them effortlessly. On the other hand, organizations make huge investments in money, time and resources to predict fraud in near real-time, if not real time and try to mitigate the consequences of fraud. Financial fraud manifests itself in various areas such as banking, insurance and investments (stock markets). It can be both offline as well as online. Online fraud includes credit/debit card fraud, transaction fraud, cyber fraud involving security, while offline fraud includes accounting fraud, forgeries etc.

Deep learning algorithms proliferated during the last five years having found immense applications in many fields, where the traditional neural networks were applied with great success. Fraud detection one of them.
In what follows, we review the works that employed deep learning for fraud detection and appeared in refereed international journals and one article is from arXive repository. papers published in International conferences are excluded.  

Pumsirirat (2018)\cite{Pumsirirat2018} proposed an unsupervised  deep auto encoder (AE) based on restricted Boltzmann machine (RBM) in order to detect novel frauds because fraudsters always try to be innovative in their modus operandi so that they are not caught while perpetrating the fraud. He employed backpropagation trained deep Auto-encoder  based on RBM  that can reconstruct normal transactions  to find anomalies  from  normal  patterns.  He used the Tensorflow library from Google to implement AE,  RBM,  and H2O  by using deep  learning.  The  results  show  the  mean  squared  error,  root mean squared error, and area under curve.

Schreyer (2017) \cite{DBLP:journals/corr/abs-1709-05254} observed the disadvantage of business and experiential-knowledge driven rules in failing to generalize well beyond the known scenarios in large scale accounting frauds. Therefore, he proposed a deep auto encoder for this purpose and tested it effectiveness on two real world datasets. Chartered accountants appreciated the power of the deep auto encoder in predicting the anomalous accounting entries.

Automobile insurance fraud has traditionally been predicted by considering only structured data and textual date present in the claims was never analyzed. But, Wang and Xu (2018) \cite{Wang2018LeveragingDL} proposed a novel method, wherein Latent  Dirichlet  Allocation  (LDA) was first used to extract the  text features  hidden in the text descriptions of the accidents appearing in the claims, and then along with the traditional  numeric  features as input data deep neural networks are trained. Based on  the  real-world  insurance fraud  dataset, they concluded their hybrid approach outperformed random forests and support vector machine.

Telecom fraud has assumed large proportions and its impact can be seen in services involving mobile banking. Zheng et al. (2018)\cite{ZHENG201878} proposed a novel generative adversarial network (GAN) based model to compute probability of fraud for each large transfer so that the bank can prevent potential frauds if the probability exceeds a threshold. The model uses a deep denoising autoencoder to learn the complex probabilistic relationship among the input features, and employs adversarial training to accurately discriminate between positive samples and negative samples in a data. They concluded that the model outperformed traditional classifiers and using it two commercial banks have reduced losses of about 10 million RMB in twelve weeks thereby significantly improving their reputation.

In today's word-of-mouth marketing, online reviews posted by customers critically influence buyers’ purchase decisions more than before. However, fraud can be perpetrated in these reviews too by posting fake and meaningless reviews, which cannot reflect customers'/users’ genuine purchase experience and opinions. They pose great challenges for users to make right choices. Therefore, it is desirable to build a fraud detection model to identify and weed out fake reviews. Dong et al. (2018)\cite{DONG2018} present an autoencoder and random forest, where a stochastic decision tree model fine tunes the parameters. Extensive experiments were conducted on a large Amazon review dataset. 

Gomez et al. (2018)\cite{GOMEZ2018175} presented a neural network based system for fraud detection in banking. They analyzed a real world dataset, and proposed an end-to-end solution from the practitioner’s perspective, especially focusing on issues such as data imbalances, data processing and cost metric evaluation. They reported  their proposed solution performed comparably with state-of-the-art  solutions.

Ryman-Tubb et al. (2018) \cite{RYMANTUBB2018130} observed that payment card fraud has dented economies to the tune of USD 416bn in 2017. This fraud is perpetrated primarily to finance terrorism, arms and drug crime. Until recently the patterns of fraud and the criminals modus operandi has remained unsophisticated. However,  smart phones, mobile payments, cloud computing and contactless payments have emerged almost simultaneously with large-scale data breaches. This made the extant methods less effective. They surveyed extant methods using transactional volumes in 2017. This benchmark will show that only eight traditional methods have a practical performance to be deployed in industry. Further, they suggested that a cognitive computing approach and deep learning are promising research directions.

Fiore et al (2019) \cite{FIORE2019448} observed that data imbalance is a crucial issue in payment card fraud detection and that oversampling has some drawbacks. They proposed Generative Adversarial Networks (GAN) for oversampling, where they trained a GAN to output mimicked minority class examples, which were then merged with training data into an augmented training set so that the effectiveness of a classifier can be improved. They concluded  that a classifier trained on the augmented set outperformed the same classifier trained on the original data, especially as far the sensitivity is concerned, resulting in an effective fraud detection mechanism.

In summary, as far as fraud detection is concerned, some progress is made in the application of a few deep learning architectures. However, there is immense potential to contribute to this field especially, the application of Resnet, gated recurrent unit, capsule network etc to detect frauds  including cyber frauds.
.
\subsection{\textbf{Financial Time Series Forecasting}} 
Advances in technology and break through in deep learning models have seen an increase in intelligent automated trading and decision support systems in Financial markets, especially in the stock and foreign exchange (FOREX) markets. However, time series problems are difficult to predict especially financial time series \cite{B1_cavalcante2016computational}. On the other hand, NN and deep learning models have shown great success in forecasting financial time series \cite{LI20097818} despite the contradictory report by efficient market hypothesis (EMH) \cite{fred12_fama1995random}, that the FOREX and stock market follows a random walk and any profit made is by chance. This can be attributed to the ability of NN to self-adapt to any nonlinear data set without any statically assumption and prior knowledge of the data set \cite{LU2009115}.

Deep leaning algorithms have used both fundamental and technical analysis data, which is the two most commonly used techniques for financial time series forecasting, to trained and build deep leaning models \cite{B1_cavalcante2016computational}. Fundamental analysis is the use or mining of textual information like financial news, company financial reports and other economic factors like government policies, to predict price movement. Technical analysis on the other hand, is the analysis of historical data of the stock and FOREX market.

Deep Learning NN (DLNN) or Multilayer Feed forward NN (MFF)  is the most used algorithms for financial markets \cite{TKAC2016788}. According to the experimental analysis done by Pandey el at \cite{PANDEY2018}, showed that MFF with Bayesian learning performed better than MFF learning with back propagation for the FOREX market.

Deep neural networks or machine learning models are considered as a black box, because the internal workings is not fully understood. The performance of DNN is highly influence by the its parameters for a particular domain. Lasfer el at \cite{lasfer2013neural} performed an analysis on the influence of parameter (like the number of neurons, learning rate, activation function etc) on stock price forecasting. The authors work showed that a larger NN produces a better result than a smaller NN. However, the effect of the activation function on a large NN is lesser.

Although CNN is well known for its stripes in image recognition and less application in the Financial markets, CNN have also shown good performance in forecasting the stock market. As indicated by \cite{lasfer2013neural}, the deeper the network the more NN can generalize to produce good results. However, the more the layers of NN, it is more likely to overfit a given data set. CNN on the other hand, with its techniques of convolution, pooling and drop out mechanism reduces the tendency of overfitting \cite{gudelek2017deep}.  

In order to apply CNN for the Financial market, the input data need to be transformed or adapted for CNN. With the help of a sliding window, Gudelek el at \cite{gudelek2017deep} used images generated by taking snapshots of the stock time series data and then fed it into 2D-CNN to perform daily predictions and classification of trends (whether downwards or upwards). The model was able to get 72 percent accuracy on 17 exchange traded fund data set. The model was not compared against other NN architecture.  
Fisher and Krauss \cite{fischer2018deep} adapted LSTM for stock prediction and compared its performance with memory-free based algorithms like random forest, logistic regression classifier and deep neural network.  LSTM performed better than other algorithms, random forest however, outperformed LSTM during the financial crisis in 2008.

EMH \cite{fred12_fama1995random} holds the view that financial news which affects the price movement are in cooperated into the price immediately or gradual. Therefore, any investor that can first analyze the news and make a good trading strategy can profit. Based on this view and the rise of big data, there has been an upward trend in sentiment analysis and text mining research which utilizes blogs, financial news social media to forecast the stock or FOREX market \cite{B1_cavalcante2016computational}. 
Santos et al \cite{dos2017stock} explored the impact of news articles on company stock prices by implementing a LSTM neural network pre-trained by a character level language model to predict the changes in prices of a company for both inter day and intraday trading. The results showed that, CNN with word wise based model outperformed other models. However, LSTM character level-based model performed better than RNN base models and also has less architectural complexity than other algorithms.

Moreover, there has been hybrid architectures to combine the strengths or more than one deep leaning models to forecast financial time series. Bao et al \cite{bao2017deep} combined wavelet transform, stacked autoencoders and LSTM for stock price prediction. The output of one network or model was fed into the next model as input. The hybrid model perfumed better than LSTM and RNN (which were standalone). Hossain et al \cite{1077}, also created a hybrid model by combining LSTM and Gated recurrent unit (GRU) to predict S\&P 500 stock price. The model was compared against standalone models like LSTM and GRU with different architectural layers. The hybrid model outperformed all other algorithms.

Calvez and Cliff \cite{Calvez2018DeepLC} did introduce a new approach on how to trade on the stock market with DLNN model. DLNN model  learn or observe the trading behaviors of traders. The author used a limit-order-book (LOB) and quotes made by successful traders (both automated and humans) as input data.  DLNN was able to learn and outperformed both human traders and automated traders. This approach of learning might be the breakthrough for intelligent automated trading for Financial markets.

\subsection {\textbf{Prognostics and Health Management}}

The service reliability of the ever-encompassing cyber-physical systems around us has started to garner the undivided attention of the prognostics community in recent years. Factors such as revenue loss, system downtime, failure in mission-critical deployments and market competitive index are emergent motivations behind making accurate predictions about the State-of-Health (SoH) and Remaining Useful Life (RUL) of components and systems. Industry niches such as manufacturing, electronics, automotive, defense and aerospace are increasingly becoming reliant on expert diagnosis of system health and smart recommender systems for maximizing system uptime and adaptive scheduling of maintenance. Given the surge in sensor influx, if there exists sufficient structured information in historical or transient data, accurate models describing the system evolution may be proposed. The general idea is that in such approaches, there is a point in the operational cycle of a component beyond which it no longer delivers optimum performance. In this regard, the most widely used metric for determining the critical operational cycle is termed as the Remaining Useful Life (RUL), which is a measure of the time from measurement to the critical cycle beyond which sub-optimal performance is anticipated. Prognostic approaches may be divided into three categorizations: (a) Model-driven (b) Data-driven (c) Hybrid i.e. any combination of (a) and (b). The last three decades have seen extensive usage of model-driven approaches with Gaussian Processes and Sequential Monte-Carlo (SMC) methods which continue to be popular in capturing patterns in relatively simpler sensor data streams. However, one shortcoming of model driven approaches used till date happens to be their dependence on physical evolution equations recommended by an expert with problem-specific domain knowledge. For model-driven approaches to continue to perform as well when the problem complexity scales, the prior distribution (physical equations) needs to continue to capture the embedded causalities in the data accurately. However, it has been the observation that as sensor data scales, the ability of model-driven approaches to learn the inherent structures in the data has lagged. This is of course due to the use of simplistic priors and updates which are unable to capture the complex functional relationships from the high dimensional input data. With the introduction of self-regulated learning paradigms such as Deep Learning, this problem of learning the structure in sensor data was mitigated to a large extent because it was no longer necessary for an expert to hand-design the physical evolution scheme of the system. With the recent advancements in parallel computational capabilities, techniques leveraging the volume of available data have begun to shine. One key issue to keep in mind is that the performance of data-driven approaches are only as good as the labeled data available for training. While the surplus of sensor data may act as a motivation for choosing such approaches, it is critical that the precursor to the supervised part of learning, i.e. data labeling is accurate. This often requires laborious and time-consuming efforts and is not guaranteed to result in the generation of near-accurate ground truth. However, when adequate precaution is in place and strategic implementation facilitating optimal learning is achieved, it is possible to deliver customized solutions to complex prediction problems with an accuracy unmatched by simpler, model-driven approaches. Therein lies the holy grail of deep learning: the ability to scale learning with training data.

The recent works on device health forecasting are as follows: Basak et al. \cite{2018arXiv181008985B} carried on Remaining Useful Life (RUL) prediction of hard disks along with discussions on effective feature normalization strategies on Backblaze hard disk data. Deep Belief Networks (DBN) based multisensor health diagnosis methodology has been proposed in \cite{TAMILSELVAN2013124} and employed in aircraft engine and electric power transformer health diagnosis to show the effectiveness of the approach.

 Kuremoto et al., \cite{KUREMOTO201447} applied DBN composed of two Restricted Botzmann Machines (RBM) to capture the input feature distribution and then optimized the size of the network and learning rate through Particle Swarm Optimization for forecasting purposes with time series data. Qiu et al., \cite{QIU20151710} proposed an early warning model where feature extraction through DNN with hidden state analysis of Hidden Markov Model (HMM) is carried out for health maintenance of equipment chain in gas pipeline.  Gugulothu et al. \cite{Gugulothu2017PredictingRU} proposed a forecasting scheme using a Recurrent Neural Network (RNN) model to generate embeddings which capture the  trend of multivariate time series data which are supposed to be disparate for healthy and unhealthy devices. The idea of using RNNs to capture intricate dependencies among various time cycles of sensor observations is emphasized in \cite{filonov} for prognostic applications. Botezatu et al., came up with some rules for directly identifying the healthy or unhealthy state of a device in \cite{Botezatu:2016:PDR:2939672.2939699}, employing a disk replacement prediction algorithm with changepoint detection applied to time series Backblaze data.
 
 So, a typical flow of prognostics and health management of any system under test using data-driven approaches start with data collection including instances of device performances or features under both normal and degraded operating conditions. The data preprocessing and feature selection play a crucial role before applying deep learning approaches to learn a model capturing degradation of device under test which should be employed to diagnose a fault, prognosticate future states of a device ensuring proper device maintenance. In this case maintaining a balance between false positives and negatives becomes crucial under real world industrial constraints and accuracy measures such as precision and recall must be validated before deployment. 

\subsection {\textbf{Medical Image Processing}} 

Deep learning techniques have pervaded the entire discipline of medical image processing and the number of studies highlighting its application in canonical tasks such as image classification, detection, enhancement, image generation, registration and segmentation have been on a sharp rise. A recent survey by Litjens et al. \cite{DBLP:journals/corr/LitjensKBSCGLGS17} presents a collective picture of the prevalence and applications of deep learning models within the community as does a fairly rigorous treatise of the same by Shen et al.  \cite{doi:10.1146/annurev-bioeng-071516-044442}. A concise overview of recent work in some of these canonical tasks follows.
 
 The purpose of image/exam classification jobs is to identify the presence of a disease based on the images of medical examinations. Over the last few years, various neural network architectures have been used in this field including stacked auto-encoders applied to diagnosis of Alzheimer’s disease and mild cognitive impairment, exploiting the latent non-linear complicated relations among various features \cite{Suk2013LatentFR}, Restricted Boltzmann Machines applied to Lung CT analysis combining generative as well as discriminative learning techniques \cite{7401039}, Deep Belief Networks trained on three dimensional medical images \cite{10.1007/978-3-642-40763-5_78} etc. Recently, the the trend of using Convolutional Neural Networks in the field of image processing has been observed. In 2017, Esteva et al. \cite{esteva2017dermatologistlevel} used and fine-tuned the Inception v3 \cite{Szegedy2016RethinkingTI} model to classify clinical images pertaining to skin cancer examinations into benign and malignant variants. Validated of experiments was carried out by testing model performance against a good number of dermatologists. In 2018, Rajaraman et. al \cite{Rajaraman2018} used specialized CNN architectures like ResNet for detecting malarial parasites in thin blood smear images. Kang et al. \cite{Kang20173DMC} improved the performance of 2D CNN by using a 3D multi-view CNN for lung nodule classification using spatial contextual information with the help of 3D Inception-ResNet architecture.
   
 Object/lesion detection aims to identify different parts/lesions in an image. Although object classification and object detection are quite similar to each other but the challenges are specific to each of the categories. When it comes to object detection, the problem of class-imbalance can pose a major hurdle in terms of the performance of object detection models. Object detection also involves identification of localized information (that is specific to different parts of an image) from the full image space. Therefore, the task of object detection  is a combination of  identification of localized information and classification \cite{6247908}. In 2016, Hwang and Kim proposed a self-transfer learning (STL) framework which optimizes both the aspects of medical object detection task. They tested the STL framework for the detection of nodules in chest radiographs and lesions in mammography \cite{HwangAndKim}.
   
   \begin{figure}
        \begin{center}
        \includegraphics[scale=0.18]{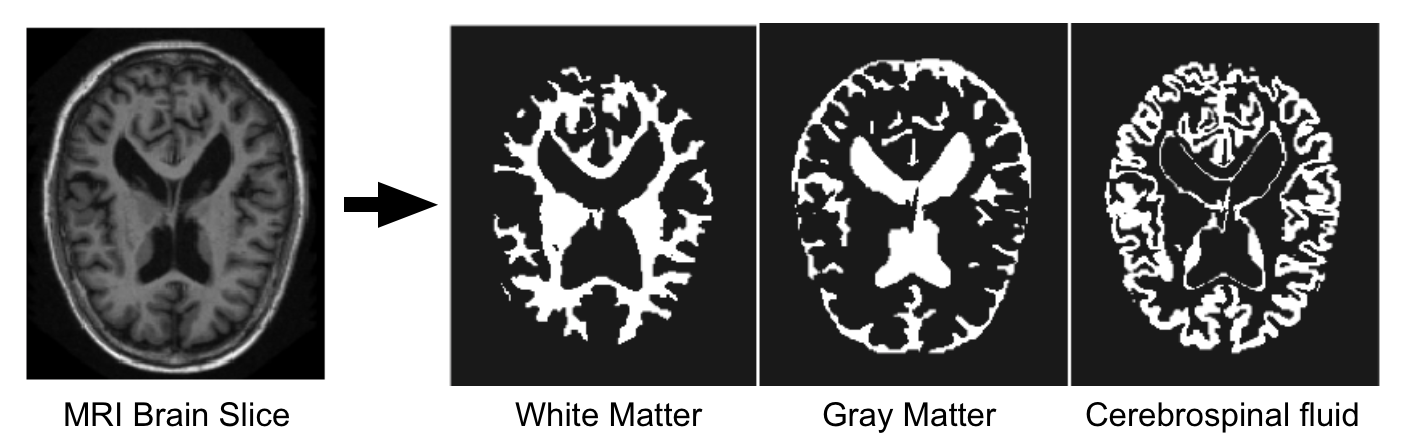}
        \caption{MRI Brain Slice and its different segmentation \cite{Segmentation}}
        \label{fig:segmentfig}
        \end{center}
   \end{figure}

   Segmentation happens to be one of the most common subjects of interest when it comes to application of Deep Learning in the domain of medical image processing. Organ and substructure segmentation allows for advanced fine-grained analysis of a medical image and it is widely practiced in the analyses of cardiac and brain images. A demonstration is shown in Figure \ref{fig:segmentfig}, where different segmented parts of an MRI Brain Slice along with the original slice are considered. Segmentation includes both the local and global context of pixels with respect to a given image and the performance of a segmentation model can suffer from inconsistencies due to class imbalances. This makes the task of segmentation a difficult one. The most widely-used CNN architecture for medical image segmentation is U-Net which was proposed by Ronneberger et al. \cite{DBLP:journals/corr/RonnebergerFB15} in 2015. U-Net takes care of sampling that is required to check the class-imbalance factors and it is capable of scanning an entire image in just one forward pass which enables it to consider the full context of the image. RNN-based architectures have also been proposed for segmentation tasks. In 2016, Andermatt et al. \cite{10.1007/978-3-319-46976-8_15} presented a method to automatically segment 3D volumes of biomedical images. They used multi-dimensional gated recurrent units (GRU) as the main layers of their neural network model. The proposed method also involves on-the-fly data augmentation which enables the model to be trained with less amount of training data.  
   
   Other applications of deep learning in Medical Image processing include image registration which implies coordinate transformation from a reference image space to target image space. Cheng et al. \cite{Cheng2018DeepSL} used multi-modal stacked denoising autoencoder to compute effective similarity measure among images using normalized mutual information and local cross correlation. On the other hand, Miao et al. \cite{7393571} developed CNN regressors to directly evaluate the registration transformation parameters. In addition to these, image generation and enhancement techniques have been discussed in \cite{Oktay2016MultiinputCI}, \cite{7448418}.
   
   So far, applications of deep learning in medical image processing has produced satisfactory results in most of the cases, However, in a sensitive field like medical image processing prior knowledge should be incorporated in cases of image detection and recognition, reconstruction so that the data driven approaches do not produce implausible results \cite{MAIER201986}. 


\subsection {\textbf{Power Systems}}
Artificial Neural Networks (ANN) have rapidly gained popularity among power system researchers \cite{vankayala1993artificial}. Since their introduction to the power systems area in 1988 \cite{chow1988incipient}, numerous applications of ANN to problems of electric power systems have been proposed. However, the recent developments of Deep Learning (DL) methods have resulted into powerful tools that can handle large data-sets and often outperform traditional machine learning methods in problems related to the power sector \cite{guo2018deep}. For this reason, currently deep architectures are receiving the attention of researchers in power industry applications. Here, we will focus on describing some approaches of deep ANN architectures applied on three basic problems of the power industry, i.e. load forecasting and  prediction of the power output of wind and solar energy systems.

Load forecasting is one of the most important tasks for the efficient power system's operation. It allows the system operator to schedule spinning reserve allocation, decide for possible interchanges with other utilities and assess system's security \cite{bourguet1994artificial}. A small decrease in load forecasting error may result in significant reduction of the total operation cost of the power system \cite{bunn1985comparative}. Among the Artificial Intelligence techniques applied for load forecasting, methods based on ANN have undoubtedly received the largest share of attention \cite{hippert2001neural}. A basic reason for their popularity lies on the fact that ANN techniques are well-suited for energy forecast \cite{kuster2017electrical}; they may obtain adequate estimations in cases where data is incomplete \cite{aggarwal1997artificial} and can consistently deal with complex non-linear problems \cite{zhai2005time}. Park et al. \cite{park1991electric}, was one of the first approaches for applying ANN in load forecasting. The efficiency of the proposed Multi-layer Perceptron (MLP) was demonstrated by benchmarking it against a numerical forecasting method frequently used by utilities. As an evolution of ANN forecasting techniques, DL methods are expected to increase the prediction accuracy by allowing higher levels of abstraction \cite{mocanu2016deep}. Thus, DL methods are gradually gain increased popularity due to their ability to capture data behaviour when considering complex non-linear patterns and large amounts of data. In \cite{chen2018short}, an end-to-end model based on deep residual neural networks is proposed for hourly load forecasting of a single day. Only raw data of past load and temperature were used as inputs of the model. Initially, the inputs of the model are processed by several fully connected layers to produce preliminary forecast. These forecasts are then passed through a deep neural network structure constructed by residual blocks. The efficiency of the proposed model was demonstrated on data-sets from the North-American Utility and ISO-NE. In \cite{bouktif2018optimal}, a Long Short Term Memory (LSTM)-based neural network has been proposed for short and medium term load forecasting. In order to optimize the effectiveness of the proposed approach, Genetic Algorithm is used to find the optimal values for the time lags and the number of layers of the LSTM model. The efficient performance of the proposed structure was verified using electricity consumption data of the France Metropolitan. Mocanu et al. \cite{mocanu2016deep} utilized two deep learning approaches based on Restricted Boltzman Machines (RBM), i.e. conditional RBM and factored conditional RBM, for single-meter residential load forecasting. The method was benchmarked against several shallow ANN architectures and a Support Vector Machine approach, demonstrating increased efficiency compared to the competing methods. Dedinec et al. \cite{dedinec2016deep} employed a Deep Belief Network (DBN) for short term load forecasting of the Former Yugoslavian Republic of Macedonia. The proposed network comprised several stacks of RBM, which were pre-trained layer-wise. Rahman et al. \cite{rahman2018predicting} proposed two models based on the architecture of Recurrent Neural Networks (RNN) aiming to predict the medium and long term electricity consumption in residential and commercial buildings with one-hour resolution. The approach has utilized a MLP in combination with a LSTM based model using an encoder-decoder architecture. A model based on LSTM-RNN framework with appliance consumption sequences for short term residential load forecasting has been proposed in \cite{kong2017short}. The researchers have showed that their method outperforms other state-of-the-art methods for load forecasting. In \cite{dong2017short} a Convolutional Neural Network (CNN) with k-means clustering has been proposed. K-means is used to partition the large amount of data into clusters, which are then used to train the networks. The method has shown improved performance compared to the case where the k-means has not been engaged. 

The utilization of DL techniques for modelling and forecasting in systems of renewable energy is progressively increasing. Since the data in such systems are inherently noisy, they may be adequately handled with ANN architectures \cite{kalogirou2001artificial}. Moreover, because renewable energy data is complicated in nature, shallow learning models may be insufficient to identify and learn the corresponding deep non-linear and non-stationary features and traits \cite{wang2017deterministic}. Among the various renewable energy sources, wind and solar energy have gained more popularity due to their potential and high availability \cite{das2018forecasting}. As a result, in recent years the research endeavours have been focused on developing DL techniques for the problems related to the deployment of the aforementioned renewable energy sources.  

Photovolatic (PV) energy has received much attention, due to its many advantages; it is abundant, inexhaustible and clean \cite{dabra2017optimization}. However, due to the chaotic and erratic nature of the weather systems, the power output of PV energy systems is intermittent, volatile and random \cite{liu2015improved}. These uncertainties may potentially degrade the real-time control performance, reduce system economics, and thus pose a great challenge for the management and operation of electric power and energy systems \cite{jang2016solar}. For these reasons, the accuracy of forecasting of PV power output plays a major role in ensuring optimum planning and modelling of PV plants. In \cite{wang2017deterministic} a deep neural network architecture is proposed for deterministic and probabilistic PV power forecasting. The deep architecture for deterministic forecasting comprises a Wavelet Transform and a deep CNN. Moreover, the probabilistic PV power forecasting model combines the deterministic model and a spine Quantile Regression (QR) technique. The method has been evaluated on historical PV power data-sets obtained from two PV farms in Belgium, exhibiting high forecasting stability and robustness. In Gensler et al. \cite{gensler2016deep},  several deep network architectures, i.e. MLP, LSTM networks, DBN and Autoencoders, have been examined with respect to their forecasting accuracy of the PV power output. The performance of the methods is validated on actual data from PV facilities in Germany. The architecture that has exhibited the best performance is the Auto-LSTM network, which combines the feature extraction ability of the Autoencoder with the forecasting ability of the LSTM. In \cite{abdel2017accurate} an LSTM-RNN is proposed for forecasting the output power of solar PV systems. In particular, the authors examine five different LSTM network architectures in order to obtain the one with the highest forecasting accuracy at the examined data-sets, which are retrieved from two cities of Egypt. The network, which provided the highest accuracy is the LSTM with memory between batches.\newline \newline
With the advantages of non-pollution, low costs and remarkable benefits of scale, wind power is considered as one of the most important sources of energy \cite{manwell2010wind}. ANN have been widely employed for processing large amounts of data obtained from data acquisition systems of wind turbines \cite{marugan2018survey}. In recent years, many approaches based on DL architectures have been proposed for the prediction of the power output of wind power systems. In \cite{wu2016probabilistic}, a deep neural network architecture is proposed for deterministic wind power forecasting, which combines CNN and LSTM networks. The results of the model are further analyzed and evaluated based on the wind power forecasting error in order to perform probabilistic forecasting. The method has been validated on data obtained from a wind farm in China; it has managed to perform better compared to other statistical methods, i.e. ARIMA and persistence method, as well as artificial intelligence based techniques in deterministic and probabilistic wind power forecasting. Wang et al. \cite{wang2017deep} proposed a wind power forecasting method based on Wavelet Transform, CNN and ensemble technique. Their method was compared with the persistence method and two shallow ANN architectures, i.e. Back-Propagation ANN (BPANN) and Support Vector Machine, on data sets collected from wind farms in China. The results validate that their method outperforms the benchmark approaches in terms of reliability, sharpness and overall skill. In \cite{wang2018deep} a DBN model in conjunction with the k-means clustering algorithm is proposed for wind power forecasting. The proposed approach demonstrated significantly increased forecasting accuracy compared to a BPANN and a Morlet wavelet neural network on data-sets obtained from a wind farm in Spain. A data-driven multi-model wind forecasting methodology with deep feature selection is proposed in \cite{feng2017data}. In particular, a two layer ensemble technique is developed; the first layer comprises multiple machine learning models, which generate individual forecasts. In the second layer a blended algorithm is utilized to merge the forecasts derived during the first stage. Numerical results validate the efficiency of the proposed methodology compared to models employing a single algorithm. Finally, in \cite{qureshi2017wind} an approach is proposed for wind power forecasting, which combines deep Autoencoders, DBN and the concept of transfer learning. The method is tested on data-sets containing  power measurement and meteorological forecast related to components of wind, obtained from wind farms in Europe. Moreover, it is compared to commonly used baseline regression models, i.e. ARIMA and Support Vector Regressor, and derives better results in terms of MAE, RMSE and SDE compared to the benchmark algorithms.




\section{Discussions}\label{conclusions}

In this paper we presented several Deep Learning architectures starting from the foundational architectures up to the recent developments covering the aspect of their modifications and evolution over time as well as applications to specific domains. We provided a list of category wise publicly available data repositories for Deep Learning practitioners in Table \ref{tab:availabledata}. We discussed the blend of swarm intelligence in Deep Learning approaches and how the influence of one enriches other when applied to real world problems. The vastly growing use of deep learning architectures specially in safety critical systems brings us to the question, how reliable the architectures are in providing decisions even in presence of adversarial scenarios. To address this, we started by giving an overview of testing neural network architectures, various methods for adversarial test generation as well as countermeasures to be adopted against adversarial examples. Next we moved on to specific applications of deep learning including Medical Imaging, Prognostics and Health Management, Applications in Financial Services, Financial Time Series Forecasting and lastly the applications in Power Systems. For each application the current research trends as well as future research directions are discussed. Table \ref{tab:reviews} lists a collection of recent reviews in different fields of Deep Learning including computer vision, forecasting, image processing, adversarial cases, autonomous vehicles, natural language processing, recommender systems and big data analytics.

\section{Conclusions and Future Work}

\label{sec:conclusion}

In conclusion, we highlight a few open areas of research and elaborate on some of the existing lines of thoughts and studies in addressing challenges that lie within.   

\vspace{1 em}
\begin{itemize}
\item \textbf{Challenges with scarcity of data:} With growing availability of data as well as powerful and distributed processing units Deep Learning architectures can be  successfully applied to major industrial problems. However, deep learning is traditionally big data driven and lacks efficiency to learn abstractions through clear verbal definitions \cite{Marcus2018DeepLA} if not trained with billions of training samples. Also the large reliance on Convolutional Neural Networks(CNNs) especially for video recognition purposes could face exponential ineffeciency leading to their demise \cite{Sabour2017DynamicRB} which can be avoided by capsules \cite{Hinton2011TransformingA} capturing critical spatial hierarchical relationships more efficiently than CNNs with lesser data requirements. To make DL work with smaller available data sets, some of the approaches in use are data augmentation, transfer learning, recursive classification techniques as well as synthetic data generation.  One shot learning \cite{Vinyals:2016:MNO:3157382.3157504} is also bringing new avenues to learn from very few training examples which has already started showing progress in language processing and image classification tasks. More generalized techniques are being developed in this domain to make DL models learn from sparse or fewer data representations is a current research thrust.

\vspace{1 em}

\item \textbf{Adopting unsupervised approaches:} A major thrust is towards combining deep learning with unsupervised learning methods. Systems developed to set their own goals \cite{Marcus2018DeepLA} and develop problem-solving approaches in its way towards exploring the environment are the future research directions surpassing supervised approaches requiring lots of data apriori. So, the thrust of AI research including Deep Learning is towards Meta Learning, i.e., learning to learn which involves automated model designing and decision making capabilities of the algorithms. It optimizes the ability to learn various tasks from fewer training data\cite{Hsu2018UnsupervisedLV}.

\vspace{1 em}

\item \textbf{Influence of cognitive meuroscience:} Inspiration drawn from cognitive neuroscience, developmental psychology to decipher human behavioral pattern are able to bring major breakthrough in applications such as enabling artificial agents learn about spatial navigation on their own which comes naturally to most living beings \cite{Banino2018VectorbasedNU}.

\vspace{1 em}

\item \textbf{Neural networks and reinforcement learning:}
Meta-modeling approaches using Reinforcement Learning(RL) are being used for designing problem specific Neural Network architectures. In \cite{Baker2016DesigningNN} the authors introduced MetaQNN, a RL based meta-modeling algorithm to automatically generate CNN architectures for image classification by using Q-learning \cite{Watkins1992} with $\epsilon$ greedy exploration. AlphaGo, the computer program built combining reinforcement learning and CNN for playing the game `Go' achieved a great success by beating human professional 'Go' players. Also deep convolutional neural networks can work as function approximators to predict `Q' values in a reinforcement learning problem. So, a major thrust of current research is on superposition 

\onecolumn

\vspace{5 em}

\newpage

\begin{centering}
\vspace{5 em}
\begin{longtable}[hbt!]{ | m{3.5 cm}| m{11 cm}| }

\caption{A Collection of Recent Reviews on Deep Learning}

\label{tab:reviews}

\\ \hline
 \centering \textbf{Topic} & \textbf{Review Papers}\\ \hline

\centering Computer Vision  & Deep learning for visual understanding: A review \cite{Guo2016DeepLF}

Deep Learning for Computer Vision: A Brief Review \cite{Voulodimos2018DeepLF}

A Survey on Deep Learning Methods for Robot Vision
\cite{RuizdelSolar2018ASO}

Deep learning for visual understanding: A review
\cite{Guo2016DeepLF}

Deep Learning Advances in Computer Vision with 3D Data: A Survey
\cite{vision3d}

Visualizations of Deep Neural Networks in Computer Vision: A Survey \cite{Seifert2017}

\\ \hline

\centering Forecasting  & Machine Learning in Financial Crisis Prediction: A Survey \cite{fin1}

Deep Learning for Time-Series Analysis
\cite{Gamboa2017DeepLF}

A Survey on Machine Learning and Statistical Techniques in Bankruptcy Prediction \cite{fin2}

Time series forecasting using artificial neural networks methodologies: A systematic review
\cite{TEALAB2018334}

A Review of Deep Learning Methods Applied on Load Forecasting
\cite{8260682}

Trends in Machine Learning Applied to Demand \& Sales Forecasting: A Review \cite{usugacadavid:hal-01881362}

A survey on retail sales forecasting and prediction in fashion markets
\cite{fin3}

Electric load forecasting: Literature survey and classification of methods
\cite{Alfares2002ElectricLF}

A review of unsupervised feature learning and deep learning for time-series modeling
\cite{LANGKVIST201411}

\\ \hline

\centering Image Processing  & A Survey on Deep Learning in Medical Image Analysis \cite{DBLP:journals/corr/LitjensKBSCGLGS17}

A Comprehensive Survey of Deep Learning for Image Captioning \cite{Hossain:2019:CSD:3303862.3295748}

Biological image analysis using deep learning-based methods: Literature review \cite{im1}

Deep learning for remote sensing image classification: A survey \cite{doi:10.1002/widm.1264}

Deep Convolutional Neural Networks for Image Classification: A Comprehensive Review
\cite{im2}

Deep Learning for Medical Image Processing: Overview, Challenges and the Future \cite{Razzak2018}

An overview of deep learning in medical imaging focusing on MRI \cite{LUNDERVOLD2019102}

Deep Learning in Medical Ultrasound Analysis: A Review \cite{LIU2019261}

\\ \hline

\centering Adversarial Cases  & Threat of Adversarial Attacks on Deep Learning in Computer Vision: A Survey \cite{Akhtar2018ThreatOA}

Adversarial Learning in Statistical Classification: A Comprehensive Review of Defenses Against Attacks \cite{DBLP:journals/corr/abs-1904-06292}

Adversarial Attacks and Defenses Against Deep Neural Networks: A Survey
\cite{OZDAG2018152}

Adversarial Machine Learning: A Literature Review \cite{ad1}

Review of Artificial Intelligence Adversarial Attack and Defense Technologies
\cite{app9050909}

A Survey of Adversarial Machine Learning in Cyber Warfare
\cite{Duddu2018ASO}

\\ \hline

\centering Autonomous Vehicles  & Planning and Decision-Making for Autonomous Vehicles \cite{auto1}

A Review of Deep Learning Methods and Applications for Unmanned Aerial Vehicles \cite{Carrio2017ARO}

MIT Autonomous Vehicle Technology Study: Large-Scale Deep Learning Based Analysis of Driver Behavior and Interaction with Automation \cite{Fridman2017MITAV}

Perception, Planning, Control, and Coordination for Autonomous Vehicles
\cite{machines5010006}

Survey of neural networks in autonomous driving \cite{auto2}

Self-Driving Cars: A Survey
\cite{Badue2019SelfDrivingCA}

\\ \hline

\centering Natural Language Processing & Recent Trends in Deep Learning Based Natural Language Processing \cite{Young2018RecentTI}

A Survey of the Usages of Deep Learning in Natural Language Processing
\cite{Otter2018ASO}

A survey on the state-of-the-art machine learning models in the context of NLP \cite{nlp1}

Inflectional Review of Deep Learning on Natural Language Processing \cite{nlp2}

Deep learning for natural language processing: advantages and challenges \cite{nlp3}
 
Deep Learning for Natural Language Processing \cite{Xie2018DeepLF}
 
\\ \hline

\centering Recommender Systems & Deep Learning based Recommender System: A Survey and New Perspectives \cite{Zhang2019DeepLB}

A Survey of Recommender Systems Based on Deep Learning \cite{8529185}

A review on deep learning for recommender systems: challenges and remedies \cite{Batmaz2018}

Deep Learning Methods on Recommender System: A Survey of State-of-the-art \cite{Betru2017DeepLM}

Deep Learning-Based Recommendation: Current Issues and Challenges \cite{re1}

A Survey and Critique of Deep Learning on Recommender Systems
\cite{Zheng2016ASA}

\\ \hline

\centering Big Data Analytics & Efficient Machine Learning for Big Data: A Review \cite{ALJARRAH201587}

A survey on deep learning for big data \cite{ZHANG2018146}

A Survey on Data Collection for Machine Learning: a Big Data - AI Integration Perspective \cite{DBLP:journals/corr/abs-1811-03402}

A survey of machine learning for big data processing \cite{Qiu2016}

Deep learning in big data Analytics: A comparative study \cite{JAN2019275}

Deep learning applications and challenges in big data analytics \cite{Najafabadi2015}

\\ \hline

\end{longtable}
\end{centering}

\vspace{2 em}

\newpage
\begin{centering}
\begin{longtable}[hbt!]{|m{1.5 cm} | m{3.5 cm}| m{10 cm}| }


\caption{A Collection of Data Repositories for Deep Learning Practitioners}

\label{tab:availabledata}

\\ \hline
 \centering \textbf{Category} & \textbf{Dataset} & \textbf{Link}\\ \hline

\centering Image Datasets  & MNIST


CIFAR-100


Caltech 101


Caltech 256


Imagenet

COIL100



STL-10


Google Open images

\hspace{1cm}



Labelme

& \url{http://yann.lecun.com/exdb/mnist/}

\url{http://www.cs.utoronto.ca/~kriz/cifar.html}

\url{http://www.vision.caltech.edu/Image_Datasets/Caltech101/}

\url{http://www.vision.caltech.edu/Image_Datasets/Caltech256/}

\url{http://www.image-net.org/}

\url{http://www1.cs.columbia.edu/CAVE/software/softlib/coil-100.php}

\url{http://www.stanford.edu/~acoates//stl10/}

\url{https://ai.googleblog.com/2016/09/introducing-open-images-dataset.html}

\url{http://labelme.csail.mit.edu/Release3.0/browserTools/php/dataset.php}

\\ \hline

\centering Speech Datasets  & Google Audioset


TIMIT

\hspace{1cm}


VoxForge

CHIME


2000 HUB5 English


LibriSpeech

VoxCeleb


Open SLR

CALLHOME American English Speech

& \url{https://research.google.com/audioset/dataset/index.html}

\url{http://www.ldc.upenn.edu/Catalog/CatalogEntry.jsp?catalogId=LDC93S1}

\url{http://www.voxforge.org/}

\url{http://spandh.dcs.shef.ac.uk/chime_challenge/data.html}

\url{https://catalog.ldc.upenn.edu/LDC2002T43}

\url{http://www.openslr.org/12/}

\url{http://www.robots.ox.ac.uk/~vgg/data/voxceleb/}

\url{https://www.openslr.org/51}

\url{https://catalog.ldc.upenn.edu/LDC97S42}

\hspace{1 cm}

\\ \hline

\centering Text Datasets  & English Broadcast News

SQuAD


Billion Word Dataset


20 Newsgroups


Google Books Ngrams


UCI Spambase


Common Crawl

Yelp Open Dataset

& \url{https://catalog.ldc.upenn.edu/LDC97S44}

\url{https://rajpurkar.github.io/SQuAD-explorer/}

\url{http://www.statmt.org/lm-benchmark/}

\url{http://qwone.com/~jason/20Newsgroups/}

\url{https://aws.amazon.com/datasets/google-books-ngrams/}

\url{https://archive.ics.uci.edu/ml/datasets/Spambase}

\url{http://commoncrawl.org/the-data/}

\url{https://www.yelp.com/dataset}

\\ \hline 

\centering Natural Language Datasets  & Web 1T 5-gram


Blizzard Challenge 2018


Flickr personal taxonomies


Multi-Domain Sentiment Dataset

Enron Email Dataset

Blogger Corpus


Wikipedia Links Data


Gutenberg eBooks List


SMS Spam Collection


UCI's Spambase data

& \url{https://catalog.ldc.upenn.edu/LDC2006T13}

\url{https://www.synsig.org/index.php/Blizzard_Challenge_2018}

\url{https://www.isi.edu/~lerman/downloads/flickr/flickr_taxonomies.html}

\hspace{1 cm}

\url{http://www.cs.jhu.edu/~mdredze/datasets/sentiment/}

\hspace{1 cm}

\url{https://www.cs.cmu.edu/~./enron/}

\url{http://u.cs.biu.ac.il/~koppel/BlogCorpus.htm}

\url{https://code.google.com/archive/p/wiki-links/downloads}

\url{http://www.gutenberg.org/wiki/Gutenberg:Offline_Catalogs}

\url{http://www.dt.fee.unicamp.br/~tiago/smsspamcollection/}

\url{https://archive.ics.uci.edu/ml/datasets/Spambase}

\\ \hline

\centering Geospatial Datasets  & OpenStreetMap

Landsat8


NEXRAD


ESRI Open data


USGS EarthExplorer

OpenTopography

NASA SEDAC


NASA Earth Observations

Terra Populus

& \url{https://www.openstreetmap.org}

\url{https://landsat.gsfc.nasa.gov/landsat-8/}

\url{https://www.ncdc.noaa.gov/data-access/radar-data/nexrad}

\url{https://hub.arcgis.com/pages/open-data}

\url{https://earthexplorer.usgs.gov/}

\url{https://opentopography.org/}

\url{https://sedac.ciesin.columbia.edu/}

\url{https://neo.sci.gsfc.nasa.gov/}

\hspace{1cm}

\url{https://terra.ipums.org/}

\\ \hline


\centering Recom- mender Systems Datasets & Movielens


Million Song Dataset


Last.fm


Book-crossing Dataset


Jester


Netflix Prize

Pinterest Fashion Compatibility


Amazon Question and Answer Data


Social Circles Data

& \url{https://grouplens.org/datasets/movielens/}

\url{https://www.kaggle.com/c/msdchallenge}


\url{https://grouplens.org/datasets/hetrec-2011/}


\url{http://www2.informatik.uni-freiburg.de/~cziegler/BX/}

\url{https://goldberg.berkeley.edu/jester-data/}

\url{https://www.netflixprize.com/}

\url{http://cseweb.ucsd.edu/~jmcauley/datasets.html#pinterest}

\hspace{1cm}

\url{http://cseweb.ucsd.edu/~jmcauley/datasets.html#amazon_qa}

\hspace{1cm}

\url{http://cseweb.ucsd.edu/~jmcauley/datasets.html#socialcircles}

\\ \hline

\centering Economics and Finance Datasets & Quandl

World Bank Open Data

IMF Data

Financial Times Market Data

Google Trends

\hspace{1cm}  


American Economic Association

US stock Data


World Factbook


Dow Jones Index Data Set

& \url{https://www.quandl.com/}

\url{https://data.worldbank.org/}

\url{https://www.imf.org/en/Data}

\url{https://markets.ft.com/data/}

\hspace{1cm}  

\url{https://trends.google.com/trends/?q=google&ctab=0&geo=all&date=all&sort=0}

\url{https://www.aeaweb.org/resources/data/us-macro-regional}

\hspace{1cm}

\url{https://github.com/eliangcs/pystock-data}

\url{https://www.cia.gov/library/publications/download/}

\url{http://archive.ics.uci.edu/ml/datasets/Dow+Jones+Index}

\hspace{1cm}

\\ \hline

\centering Auto- nomous Vehicles Datasets & BDD100k

Baidu Apolloscapes

Comma.ai


Oxford’s Robotic Car


Cityscape Dataset


CSSAD Dataset


KUL Belgium Traffic Sign Dataset

LISA


Bosch Small Traffic Light 

LaRa Traffic Light Recognition


WPI Datasets

& \url{https://bdd-data.berkeley.edu/}

\url{http://apolloscape.auto/}

\url{https://archive.org/details/comma-dataset}

\url{https://robotcar-dataset.robots.ox.ac.uk/}

\url{https://www.cityscapes-dataset.com/}

\url{http://aplicaciones.cimat.mx/Personal/jbhayet/ccsad-dataset}

\url{http://www.vision.ee.ethz.ch/~timofter/traffic_signs/}

\hspace{1cm}

\url{http://cvrr.ucsd.edu/LISA/datasets.html}

\url{https://hci.iwr.uni-heidelberg.de/node/6132}

\hspace{1cm}

\url{http://www.lara.prd.fr/benchmarks/trafficlightsrecognition}

\hspace{1cm}

\url{http://computing.wpi.edu/dataset.html}

\label{tab:repository}
\\ \hline
\end{longtable}
\end{centering}

  
\vspace{10 cm}
  

\twocolumn

\vspace{1 em}
of neural networks and reinforcement learning geared towards problem specific requirements.
\end{itemize} 
\vspace{1 em}
This review has aimed at aiding the beginner as well as the practitioner in the field make informed choices and has made an in-depth analysis of some recent deep learning architectures as well as an exploratory dissection of some pertinent application areas. It is the authors' hope that readers find the material engaging and informative and openly encourage feedback to make the organization and content of this article more aligned along the lines of a formal extension of the literature within the deep learning community.



\bibliographystyle{IEEEtran}
\bibliography{IEEEabrv,bibliography}

\vfill


\end{document}